\documentclass[11pt,a4paper]{article}
\usepackage[hyperref]{emnlp2020}
\usepackage{times}
\usepackage{latexsym}

\usepackage{microtype}

\aclfinalcopy 
\def\aclpaperid{183} 

\title{GRUEN for Evaluating Linguistic Quality of Generated Text}

\author{Wanzheng Zhu \and Suma Bhat\\
  University of Illinois at Urbana-Champaign, USA \\
  \texttt{wz6@illinois.edu}, \texttt{spbhat2@illinois.edu} \\ 
}

\date{}
\usepackage{graphicx, amsmath, amsfonts, amssymb, xspace, color, enumitem}
\usepackage{amsthm}
\usepackage{amsfonts}
\usepackage{xcolor}
\usepackage{booktabs}
\usepackage{tabularx} 
\usepackage[noend,ruled,linesnumbered]{algorithm2e}
\usepackage{multirow}

\newcommand{\ie}{\emph{i.e.}\xspace} 
\newcommand{\eg}{\emph{e.g.}\xspace} 
\newcommand{\our}{\textsc{GRUEN}\xspace}
\newcommand{\nop}[1]{}

\begin{document}
\maketitle
\begin{abstract}
	Automatic evaluation metrics are indispensable for evaluating generated text. 
	To date, these metrics have focused almost exclusively on the content selection aspect of the system output, ignoring the linguistic quality aspect altogether. 
	We bridge this gap by proposing \our for evaluating Grammaticality, non-Redundancy, focUs, structure and coherENce of generated text.\footnote{Following BLEU and ROUGE -- blue and red in French, we name our evaluation metric \our \xspace -- that means green in German.} 
	\our utilizes a BERT-based model and a class of syntactic, semantic, and contextual features to examine the system output. 
	Unlike most existing evaluation metrics which require human references as an input, \our is reference-less and requires only the system output. 
	Besides, it has the advantage of being unsupervised, deterministic, and adaptable to various tasks. 
	Experiments on seven datasets over four language generation tasks show that the proposed metric correlates highly with human judgments.\footnote{Our metric is available at \url{https://github.com/WanzhengZhu/GRUEN}.}
\end{abstract}

\section{Introduction}
\label{sec:intro}

Automatic evaluation metrics for Natural Language Generation (NLG) tasks reduce the need for human evaluations, which can be expensive and time-consuming to collect. Fully automatic metrics allow faster measures of progress when training and testing models, and therefore, accelerate the development of NLG systems \cite{chaganty2018price,zhang2020bertscore,clark2019sentence}.

\begin{table}[ht]
	\centering
	\small
	\begin{tabular}{p{0.46\textwidth}}
		\toprule
		\textbf{Q1: Grammaticality} The summary should have no datelines, system-internal formatting, capitalization errors or obviously ungrammatical sentences (\eg, fragments, missing components) that make the text difficult to read.\\
		\textbf{Q2: Non-redundancy} There should be no unnecessary repetition in the summary.\\ 
		\textbf{Q3: Focus} The summary should have a focus; sentences should only contain information that is related to the rest of the summary.\\
		\textbf{Q4: Structure and Coherence} The summary should be well-structured and well-organized. The summary should not just be a heap of related information, but should build from sentence to sentence to a coherent body of information about a topic.\\
		\bottomrule
	\end{tabular}
	\caption{Dimensions of linguistic quality as proposed in \citet{hoa2006overview}. }
	\label{table:DUC_guideline}
\end{table}

To date, most automatic metrics have focused on measuring the content selection between the human references and the model output, leaving  linguistic quality to be only indirectly captured (\eg, n-gram and longest common subsequence in ROUGE-N and ROUGE-L respectively \cite{lin2003automatic,lin2004rouge}, and alignment in METEOR \cite{banerjee2005meteor}). 
Even though the need for an explicit measure of linguistic quality has long been pointed out in \citet{hoa2006overview,conroy2008mind}, this aspect has remained under-explored barring a few studies that focused on measuring the linguistic quality of a generated piece of text \cite{pitler2010automatic,kate2010learning,xenouleas2019sum}.

In this paper, we bridge this gap by proposing a novel metric for evaluating the \textit{linguistic quality} of system output.
Taking into consideration the guidelines put forth for the Document Understanding Conference (DUC) in Table \ref{table:DUC_guideline}, we evaluate:
1) \textit{Grammaticality} by computing the sentence likelihood and the grammatical acceptability with a BERT-based language representation model \cite{devlin2019bert}, 
2) \textit{Non-redundancy} by identifying repeated components with inter-sentence syntactic features, 
3) \textit{Focus} by examining semantic relatedness between adjacent sentences using Word Mover's Distance (WMD) \cite{kusner2015word}, and 
4) \textit{Structure and Coherence} by measuring the Sentence-Order Prediction (SOP) loss with A Lite BERT \cite{lan2019albert}.

Compared with existing metrics, \our is advantageous in that it is: 
\begin{itemize}[leftmargin=*]
	\item \textit{Most correlated} with human judgments: It achieves the highest correlation with human judgments when compared with other metrics of linguistic quality, demonstrated using seven datasets over four NLG tasks. 
	\item \textit{Reference-less}: Most existing evaluation metrics (\eg, ROUGE, METEOR, MoverScore \cite{zhao2019moverscore}) require human references for comparison. However, it is only logical to assume that the linguistic quality of a system output should be measurable from the output alone.
	To that end, \our is designed to be reference-less, and requires only the system output as its input. 
	\item \textit{Unsupervised}: Available supervised metrics (\eg, SUM-QE \cite{xenouleas2019sum}) not only require costly human judgments\footnote{We use ``human references" to mean the ground truth output for a given task, and ``human judgments" as the manual linguistic quality annotation of a system's output.} as supervision for each dataset, but also risk poor generalization to new datasets. In addition, they are non-deterministic due to the  randomness in the training process. In contrast, \our is unsupervised, free from training and deterministic. 
	\item \textit{General}: Almost all existing metrics for evaluating the linguistic quality are task-specific (\eg, \citet{pitler2010automatic} and SUM-QE \cite{xenouleas2019sum} are for text summarization), whereas  \our is more generally applicable and  performs well in various NLG task settings as we demonstrate empirically. 
\end{itemize}

\section{Related Work}
\label{sec:related_work}
The growing interest in NLG has given rise to better automatic evaluation metrics to measure the output quality.
We first review the widely used metrics for NLG tasks and then discuss available metrics for evaluating linguistic quality.

\subsection{NLG Evaluation Metrics}
\noindent \textbf{N-gram-based metrics}: 
BLEU \cite{papineni2002bleu}, ROUGE \cite{lin2003automatic,lin2004rouge} and METEOR \cite{banerjee2005meteor,lavie2009meteor,denkowski2014meteor} are three most commonly used metrics to measure the n-gram lexical overlap between the human references and the system output in various NLG tasks. 
To tackle their intrinsic shortcomings (\eg,  inability to capture lexical similarities), many variations have been proposed such as NIST \cite{doddington2002automatic},  ROUGE-WE \cite{ng2015better}, ROUGE-G \cite{shafieibavani2018graph} and METEOR++ 2.0 \cite{guo2019meteor++}.

\noindent \textbf{Embedding-based metrics}: 
These metrics utilize neural models to learn dense representations of words  \cite{mikolov2013distributed,pennington2014glove} and sentences \cite{ng2015better,pagliardini2018unsupervised,clark2019sentence}. 
Then, the embedding distances of the human references and the system output are measured by cosine similarity or Word Mover’s Distance (WMD) \cite{kusner2015word}. 
Among them, MoverScore \cite{zhao2019moverscore}, averaging n-gram embeddings with inverse document frequency, shows robust performance on different NLG tasks.

\noindent \textbf{Supervised metrics}: 
More recently, various supervised metrics have been proposed. They are trained to optimize the correlation with human judgments in the training set. 
BLEND \cite{ma2017blend} uses regression to combine various existing metrics. 
RUSE \cite{shimanaka2018ruse} leverages pre-trained sentence embedding models. 
SUM-QE \cite{xenouleas2019sum} encodes the system output by a BERT encoder and then adopts a linear regression model. 
However, all these supervised metrics not only require costly human judgments for each dataset as input, but also have the risk of poor generalization to new datasets and new domains \cite{chaganty2018price,zhang2020bertscore}. 
In contrast, unsupervised metrics require no additional human judgments for new datasets or tasks, and can be generally used for various datasets/tasks.

\noindent \textbf{Task-specific metrics}: 
Some metrics are proposed to measure the specific aspects of the tasks. 
For instance, in text simplification, SARI \cite{xu2016optimizing} measures the simplicity gain in the output. 
In text summarization, most metrics are designed to evaluate the content selection, such as Pyramid \cite{nenkova2004evaluating}, SUPERT \cite{gao2020supert} and \citet{mao2020facet}. 
In dialogue systems, diversity and coherence are assessed in \citet{li2016diversity,li2016deep} and \citet{dziri2019evaluating}.
However, these proposed metrics are not generally applicable to the evaluation of other aspects or tasks.

\subsection{Evaluating Linguistic Quality}
\label{sec:related_work_2}
Existing metrics have focused mostly on evaluating the aspect of content selection in the system output, while ignoring the aspect of linguistic quality. 
This suggests the long-standing need for automatic measures of linguistic quality of NLG output, despite requests for further studies in this important direction. 
For instance, the Text Analysis Conference (TAC)\footnote{\url{http://tac.nist.gov/}} and the Document Understanding Conference (DUC)\footnote{\url{http://duc.nist.gov/}} \cite{hoa2006overview} have motivated the need to automatically evaluate the linguistic quality of summarization since 2006. 
As another example, \citet{conroy2008mind} have highlighted the downsides of ignoring linguistic quality while focusing on summary content during system evaluation. 
Additionally, the need for linguistic quality evaluation has been underscored in \citet{dorr2011machine,graham2013continuous,novikova2017we,way2018quality,specia2018machine}. 
The uniqueness of our study is that it bridges the need of an automatic evaluation metric of language quality to enable a more holistic evaluation of language generation systems.

Among the few existing metrics of linguistic quality available in prior studies, the early ones \citet{pitler2010automatic,kate2010learning} rely only on shallow syntactic linguistic features, such as part-of-speech tags, n-grams and named entities. 
To better represent the generated output, the recent SUM-QE model \cite{xenouleas2019sum} encodes the system output by a BERT encoder and then adopts a linear regression model to predict the linguistic quality. 
It shows the state-of-the-art results and is most relevant to our work. 
However, SUM-QE is a supervised metric, which not only requires costly human judgments as input for each dataset, but also has non-deterministic results due to the intrinsic randomness in the training process. 
Besides, SUM-QE has been shown to work well with the DUC datasets of the summarization task only \cite{xenouleas2019sum}, calling into question its effectiveness for other datasets and tasks. 
\our, as an unsupervised metric, requires no additional human judgments for new datasets and has been shown to be effective on seven datasets over four NLG tasks.

\section{Proposed Metric}
\label{sec:model}

In this section, we describe the  proposed linguistic quality metric in detail. 
We define the problem as follows: given a system output $S$ with $n$ sentences $[s_1, s_2, ..., s_n]$, where $s_i$ is any one sentence (potentially among many), we aim to output a holistic score,  $Y_S$, of its linguistic quality. 
We explicitly assess system output for the four aspects in Table \ref{table:DUC_guideline} -- Grammaticality, Non-redundancy, Focus, and Structure and Coherence. 
We leave Referential Clarity as suggested in \citet{hoa2006overview} for future work.

\noindent \textbf{Grammaticality}: 
A system output with a high grammaticality score $y_g$ is expected to be  readable, fluent and grammatically correct. 
Most existing works measure the sentence likelihood (or perplexity) with a language model. 
We, in addition, explicitly capture whether the sentence is grammatically ``acceptable" or not. 

We measure $y_g$ using two features: sentence likelihood and grammar acceptance. 
For a system output $S$, we first use the Punkt sentence tokenizer \cite{kiss2006unsupervised} to extract its component sentences $s_1, s_2, ..., s_n$. 
Then, for each  sentence $s_i = (w_{i,1}, w_{i,2}, ..., w_{i,k})$, a sequence of words $w_{i,j}$, we measure its sentence likelihood score $l_i$ and grammar acceptance score $g_i$ by a BERT model \cite{devlin2019bert}.\footnote{We use the ``bert-base-cased" model from: \\ \url{http://huggingface.co/transformers/pretrained\_models.html}.}
The choice of BERT is to leverage the contextual features and the masked language model (MLM), which can best examine the word choice. 
However, BERT can not be directly applied to get the likelihood of a sentence, as it is designed to get the probability of a single missing word. 
Inspired by \citet{wang2019bert,wang2019language}, we estimate $l_i$ by a unigram approximation of the words in the sentence: 
$l_i = \sum_{j} \log p(w_{i,j} | w_{i,1} ..., w_{i,j-1}, w_{i,j+1}, ..., w_{i,k})$. 
By such approximation, $l_i$ can be estimated by computing the masked probability of each word. 
To obtain the grammar acceptance score $g_i$, we fine-tune the BERT model on the Corpus of Linguistic Acceptability (CoLA) \cite{warstadt2018neural}, a dataset with 10,657 English sentences labeled as grammatical or ungrammatical from linguistics publications. 
Finally, scores from both models (\ie, $l_i$ and $g_i$) are linearly combined to examine the grammaticality of the sentence $s_i$. 
The final grammaticality score $y_g$ is obtained by averaging scores of all $n$ component sentences: $y_g = \sum_{i} (l_i+g_i) /n$.

\noindent \textbf{Non-redundancy}: 
As shown in \citet{hoa2006overview}, non-redundancy refers to having no unnecessary repetition, which takes the form of whole sentences or sentence fragments or noun phrases (\eg, ``Bill Clinton") when a pronoun (``he") would suffice across sentences. 
To calculate the  non-redundancy score $y_r$, we capture repeated components by using four \textit{inter-sentence} syntactic features: 1) string length of the longest common substring, 2) word count of longest common words, 3) edit distance, and 4) number of common words. 
We compute the four features for each pair of component sentences and there are  $\binom{n}{2}$ such pairs in total. 
For each pair of  sentences $(s_i, s_j)$, we count the number of times $m_{i,j}$ that these pairs are beyond a non-redundancy penalty threshold.
The penalty threshold for each feature are: 
$<$80\% string length of the shorter sentence, 
$<$80\% word count of the shorter sentence, 
$>$60\% string length of the longer sentence, and 
$<$80\% word count of the shorter sentence, respectively.
Finally, we get $y_r = -0.1* \sum_{i,j}{m_{i,j}}$. 
Note that the non-redundancy penalty threshold and penalty weight are learned empirically from a held-out validation set. 
We discuss the effectiveness of each feature in detail in Appendix \ref{sec:analysis_nonredun}.

\noindent \textbf{Focus}: 
Discourse focus has been widely studied and many phenomena show that a focused output should have related semantics between adjacent sentences \cite{walker1998centering,knott2001beyond,pitler2010automatic}. 
We compute the focus score $y_f$ by calculating semantic relatedness for each pair of adjacent sentences $(s_i, s_{i+1})$. 
Specifically, we calculate the Word Mover Similarity $wms(s_i, s_{i+1})$ \cite{kusner2015word} for the sentence pair $(s_i, s_{i+1})$. 
If the similarity score is less than the similarity threshold 0.05, we will impose a penalty score -0.1 on the focus score $y_f$. 
A focused output should expect $y_f = 0$.

\noindent \textbf{Structure and coherence}: 
A well-structured and coherent output should contain well-organized sentences, where the sentence order is natural and easy-to-follow. 
We compute the inter-sentence coherence score $y_c$ by a self-supervised loss that focuses on modeling inter-sentence coherence, namely Sentence-Order Prediction (SOP) loss. 
The SOP loss, proposed by \citet{lan2019albert}, has been shown to be more effective than the Next Sentence Prediction (NSP) loss in the original BERT \cite{devlin2019bert}. 
We formulate the SOP loss calculation as follows. 
First, for a system output $S$, we extract all possible consecutive pairs of segments (\ie, $([s_1, ..., s_i], [s_{i+1}, ..., s_n])$, where $i \in [1,2,...,n-1]$). 
Then, we take as positive examples two consecutive segments, and as negative examples the same two consecutive segments but with their order swapped. 
Finally, the SOP loss is calculated as the average of the logistic loss for all segments,\footnote{We select as the model architecture the pre-trained ALBERT-base model from \url{https://github.com/google-research/ALBERT}.} and the coherence score $y_c$ is the additive inverse number of the SOP loss.

\noindent \textbf{Final score}: 
The final linguistic quality score $Y_S$ is a linear combination of the above four scores: $Y_S = y_g + y_r + y_f + y_c$. 
Note that the final score $Y_S$ is on a scale of 0 to 1, and all the hyper-parameters are learned to maximize the Spearman's correlation with human judgments for the held-out validation set.

\section{Empirical Evaluation}
\label{sec:exp}

In this section, we evaluate the quality of different metrics on four NLG tasks: 
1) \textit{abstractive text summarization}, 2) \textit{dialogue system}, 3) \textit{text simplification} and 4) \textit{text compression}. 

\noindent \textbf{Evaluating the metrics}: 
We assess the performance of an evaluation metric by analyzing how well it correlates with human judgments. 
We, following existing literature, report Spearman's correlation $\rho$, Kendall's correlation $\tau$, and Pearson's correlation $r$.
In addition, to tackle the correlation non-independence issue (two dependent correlations sharing one variable) \cite{graham2014testing}, we report William's significance test \cite{williams1959regression}, which can reveal whether one metric significantly outperforms the other.

\noindent \textbf{Correlation type}: 
Existing automatic metrics tend to correlate poorly with human judgments at the instance-level, although several metrics have been found to have high system-level correlations \cite{chaganty2018price,novikova2017we,liu2016not}. 
Instance-level correlation is critical in the sense that error analysis can be done more constructively and effectively. 
In our paper, we primarily analyze the instance-level correlations and briefly discuss the system-level correlations.

\noindent \textbf{Baselines}: 
We compare \our with the following baselines: 
\begin{itemize}[leftmargin=*]
	\setlength\itemsep{-0.3em}
	\item \textbf{BLEU-best} \cite{papineni2002bleu} (best of BLEU-N. It refers to the version that achieves best correlations and is different across datasets.)
	\item \textbf{ROUGE-best} \cite{lin2004rouge} (best of ROUGE-N, ROUGE-L, ROUGE-W)
	\item \textbf{METEOR} \cite{lavie2009meteor}
	\item Translation Error Rate (\textbf{TER}) \cite{snover2006study}
	\item \textbf{VecSim} \cite{pagliardini2018unsupervised}
	\item \textbf{WMD-best} (best of Word Mover Distance \cite{kusner2015word}, Sentence Mover Distance \cite{clark2019sentence}, Sentence+Word Mover Distance \cite{clark2019sentence})
	\item \textbf{MoverScore} \cite{zhao2019moverscore}
	\item \textbf{SUM-QE} \cite{xenouleas2019sum} (we use the ``BERT-FT-M-1" model trained on the DUC-2006 \cite{hoa2006overview} and DUC-2007 \cite{over2007duc} datasets)
	\item \textbf{SARI} \cite{xu2016optimizing} (compared in the text simplification task only)
\end{itemize}
Note that we do not include \citet{pitler2010automatic} and \citet{kate2010learning}, since their metrics rely only on shallow syntactic linguistic features and should probably have no better results than SUM-QE \cite{xenouleas2019sum}. 
Besides, their implementations are not publicly available. 
For the complete results of BLEU, ROUGE and WMD, please refer to Table \ref{table:full_simplification}-\ref{table:full_dialogue} in Appendix.

\subsection{Abstractive Text Summarization}
\label{sec:summarization}

\begin{table*}[t!]
	\centering
	\small
	\begin{tabular}{ccc|cc|cc|cc|cc}
		\toprule
		& \multicolumn{8}{c}{\textbf{CNN/Daily Mail}} & \multicolumn{2}{|c}{\textbf{TAC-2011}} \\ 
		\cmidrule{2-11}
		& \multicolumn{2}{c}{\textbf{Overall}} & \multicolumn{2}{|c}{\textbf{Grammar}} & \multicolumn{2}{|c}{\textbf{Non-redun}} & \multicolumn{2}{|c}{\textbf{Post-edits}} & \multicolumn{2}{|c}{\textbf{Readability}} \\ 
		\cmidrule{2-11} 
		& $\rho$  & $r$ & $\rho$ &  $r$ & $\rho$ &  $r$ & $\rho$ & $r$ & $\rho$  & $r$ \\
		\midrule
		\textbf{BLEU-best} & 0.17 & 0.18 & 0.11 & 0.12 & 0.17 & 0.20 & -0.21 & -0.29 & 0.26 & 0.38 \\
		\cmidrule{2-11}
		\textbf{ROUGE-best} & 0.17 & 0.19 & 0.11 & 0.13 & 0.20 & 0.23 & -0.24 & -0.32 & 0.25 & 0.36 \\		\cmidrule{2-11}
		\textbf{METEOR} & 0.17 & 0.18 & 0.10 & 0.12 & 0.20 & 0.22 & -0.25 & -0.28 & 0.24 & 0.32 \\
		\cmidrule{2-11}
		\textbf{TER} & -0.04 & -0.03 & 0.03 & 0.02 & -0.07 & -0.08 & 0.08 & 0.08 & 0.21 & 0.34 \\
		\cmidrule{2-11}
		\textbf{VecSim} & 0.16 & 0.19 & 0.09 & 0.12 & 0.18 & 0.22 & -0.24 & -0.34 & 0.16 & 0.33 \\
		\cmidrule{2-11}
		\textbf{WMD-best} & 0.26 & 0.24 & 0.20 & 0.21 & 0.26 & 0.23 & -0.29 & -0.26 & 0.15 & 0.25 \\
		\cmidrule{2-11}
		\textbf{MoverScore} & 0.24 & 0.26 & 0.15 & 0.17 & 0.28 & 0.32 & -0.32 & -0.40 & 0.29 & 0.40 \\
		\cmidrule{2-11}
		\textbf{SUM-QE} & 0.46 & 0.48 & 0.41 & \textbf{0.41} & 0.45 & 0.44 & -0.51 & -0.43 & \textbf{0.40} & 0.41 \\
		\cmidrule{2-11}
		\textbf{\our} & \textbf{0.52} & \textbf{0.54} & \textbf{0.43} & 0.40 & \textbf{0.52} & \textbf{0.58} & \textbf{-0.60} & \textbf{-0.58} & \textbf{0.40} & \textbf{0.45} \\	
		\bottomrule
	\end{tabular}
	\caption{Instance-level Spearman's $\rho$ and Pearson's $r$ correlations on the CNN/Daily Mail and TAC-2011 datasets.}
	\label{table:summarization}
\end{table*}

\textbf{Dataset}: We evaluate \our for Text Summarization using two benchmark datasets: the \textit{CNN/Daily Mail} dataset \cite{hermann2015teaching,nallapati2016abstractive} and the \textit{TAC-2011} dataset\footnote{\url{http://tac.nist.gov/}}. 

The \textit{CNN/Daily Mail} dataset contains online news articles paired with multi-sentence summaries (3.75 sentences or 56 tokens on average). 
We obtain the human annotated linguistic quality scores from \citet{chaganty2018price} and use the 2,086 system outputs from 4 neural models. 
Each system output has human judgments on a scale from 1-3 for: \textit{Grammar}, \textit{Non-redundancy} and \textit{Overall} linguistic quality of the summary using the guideline from the DUC summarization challenge \cite{hoa2006overview}. 
In addition, it measures the number of \textit{Post-edits} to improve the summary quality. 
For all human judgments except \textit{Post-edits}, higher scores indicate better quality. 

The \textit{TAC-2011} dataset, from the Text Analysis Conference (TAC), contains 4488 data instances (4.43 sentences or 94 tokens on average). It has 88 document sets and each document set includes 4 human reference summaries and 51 summarizers. 
We report correlation results on the \textit{Readability} score, which measures the linguistic quality according to the guideline in \citet{hoa2006overview}.

\noindent \textbf{Results}: 
Instance-level correlation scores are summarized in Table~\ref{table:summarization}. 
As expected, all the baseline approaches except SUM-QE perform poorly because they do not aim to measure linguistic quality explicitly. 
We note that most of the baselines are highly unstable (and not robust) across the different datasets. 
For instance, BLEU performs relatively well on TAC-2011 but poor on CNN/Daily Mail, while WMD performs relatively well on CNN/Daily Mail but poor on TAC-2011. 
\our outperforms SUM-QE on all aspects except the Grammar of CNN/Daily Mail, where they have comparable performance. 
We performed a set of William's tests for the significance of the differences in performance between \our and SUM-QE for each linguistic score and each correlation type. 
We found that the differences were significant ($p < 0.01$) in all cases expect the Grammar of CNN/Daily Mail, as shown in Table \ref{table:william_significance_test} in Appendix. 

\subsection{Dialogue System}
\label{sec:dialogue}

\begin{table*}[t!]
	\centering
	\small
	\begin{tabular}{ccc|cc|cc|cc|cc|cc}
		\toprule
		& \multicolumn{4}{c}{\textbf{BAGEL}} & \multicolumn{4}{|c}{\textbf{SFHOTEL}} & \multicolumn{4}{|c}{\textbf{SFREST}}\\ 
		\cmidrule{2-13}
		& \multicolumn{2}{c}{\textbf{Naturalness}} & \multicolumn{2}{|c}{\textbf{Quality}} & \multicolumn{2}{|c}{\textbf{Naturalness}} & \multicolumn{2}{|c}{\textbf{Quality}} & \multicolumn{2}{|c}{\textbf{Naturalness}} & \multicolumn{2}{|c}{\textbf{Quality}} \\ 
		\cmidrule{2-13} 
		& $\rho$  & $r$ & $\rho$  & $r$ & $\rho$ & $r$ & $\rho$ & $r$ & $\rho$ & $r$ & $\rho$  & $r$\\
		\midrule
		\textbf{BLEU-best} & 0.03 & 0.04 & 0.02 & 0.05 & 0.00 & 0.07 & -0.10 & -0.02 & 0.03 & 0.03 & -0.03 & -0.02 \\
		\cmidrule{2-13}
		\textbf{ROUGE-best} & 0.11 & 0.13 & 0.10 & 0.12 & -0.02 & 0.02 & -0.12 & -0.07 & 0.02 & 0.03 & -0.06 & -0.04 \\
		\cmidrule{2-13}
		\textbf{METEOR} & 0.02 & 0.03 & 0.05 & 0.05 & -0.04 & 0.02 & -0.14 & -0.07 & 0.03 & 0.04 & -0.01 & 0.00 \\
		\cmidrule{2-13}
		\textbf{TER} & 0.11 & 0.15 & 0.11 & 0.15 & -0.01 & -0.02 & -0.05 & -0.03 & 0.01 & -0.01 & -0.06 & -0.08 \\
		\cmidrule{2-13}
		\textbf{VecSim} & 0.03 & 0.05 & 0.05 & 0.07 & -0.03 & 0.04 & -0.15 & -0.06 & 0.02 & 0.02 & -0.05 & -0.05 \\
		\cmidrule{2-13}
		\textbf{WMD-best} & 0.03 & 0.05 & 0.05 & 0.08 & -0.02 & 0.00 & -0.12 & -0.07 & 0.03 & 0.05 & -0.05 & 0.00 \\
		\cmidrule{2-13}
		\textbf{MoverScore} & 0.07 & 0.10 & 0.06 & 0.10 & -0.03 & 0.02 & -0.12 & -0.06 & 0.02 & 0.02 & -0.04 & -0.02 \\
		\cmidrule{2-13}
		\textbf{SumQE} & 0.14 & 0.17 & 0.13 & 0.16 & 0.23 & 0.30 & 0.16 & 0.24 & 0.09 & 0.11 & 0.11 & 0.13 \\
		\cmidrule{2-13}
		\textbf{\our} & \textbf{0.22} & \textbf{0.32} & \textbf{0.19} & \textbf{0.26} & \textbf{0.44} & \textbf{0.48} & \textbf{0.44} & \textbf{0.51} & \textbf{0.24} & \textbf{0.25} & \textbf{0.27} & \textbf{0.27} \\
		\bottomrule
	\end{tabular}
	\caption{Instance-level Spearman's $\rho$ and Pearson's $r$ correlations on the BAGEL, SFHOTEL and SFREST datasets.}
	\label{table:dialogue}
\end{table*}

\textbf{Dataset}: We use three task-oriented dialogue system datasets: BAGEL \cite{mairesse2010phrase}, SFHOTEL \cite{wen2015semantically} and SFREST \cite{wen2015semantically}, which contains 404, 875 and 1181 instances respectively. 
Each system output receives \textit{Naturalness} and \textit{Quality} scores \cite{novikova2017we}. 
\textit{Naturalness} measures how likely a system utterance is generated by native speakers. 
\textit{Quality} measures how well a system utterance captures fluency and grammar.

\noindent \textbf{Results} (Table \ref{table:dialogue}): \our outperforms all other metrics by a significant margin. 
Interestingly, no metric except \our produces even a moderate correlation with human judgments, regardless of dataset or aspect of human judgments. 
The finding agrees with the observations in \citet{wen2015semantically,novikova2017we,zhao2019moverscore}, where \citet{novikova2017we} attributes the poor correlation to the unbalanced label distribution. 
Moreover, we analyze the results further in Appendix \ref{sec:app_dialogue} in an attempt to interpret them.

\subsection{Text Simplification}
\label{sec:simplification}

\begin{table}[t!]
	\centering
	\small
	\begin{tabular}{cccc}
		\toprule
		& $\rho$ & $\tau$ & $r$ \\
		\midrule
		\textbf{BLEU-best} & 0.55 & 0.40 & 0.58 \\
		\cmidrule{2-4}
		\textbf{ROUGE-best} & 0.61 & 0.45 & 0.64 \\
		\cmidrule{2-4}
		\textbf{METEOR} & 0.63 & 0.47 & \textbf{0.67} \\
		\cmidrule{2-4}
		\textbf{TER} & 0.55 & 0.40 & 0.56 \\
		\cmidrule{2-4}
		\textbf{VecSim} & 0.47 & 0.34 & 0.53 \\		
		\cmidrule{2-4}
		\textbf{WMD-best} & 0.43 & 0.31 & 0.33 \\
		\cmidrule{2-4}
		\textbf{MoverScore} & 0.62 & 0.46 & 0.65 \\
		\cmidrule{2-4}
		\textbf{SumQE} & 0.62 & 0.45 & 0.64 \\
		\cmidrule{2-4}
		\textbf{SARI} & 0.35 & 0.25 & 0.40 \\
		\cmidrule{2-4}
		\textbf{\our} & \textbf{0.65} & \textbf{0.49} & 0.65 \\
		\bottomrule
	\end{tabular}
	\caption{Instance-level Spearman's $\rho$, Kendall's $\tau$ and Pearson's $r$ correlations with \textit{Grammar} on the text simplification dataset \cite{xu2016optimizing}.}
	\label{table:simplification}
\end{table}

\textbf{Dataset}: We use a benchmark text simplification dataset with 350 data instances, where each instance has one system output and eight human references \cite{xu2016optimizing}. 
Each system output instance receives a human-assigned \textit{Grammar} score. 

\noindent \textbf{Results}: Table~\ref{table:simplification} presents the results on the dataset of \citet{xu2016optimizing}. 
We note that both \our and METEOR have the best results.
The rest of the baseline metrics have satisfactory results too, such as MoverScore and ROUGE. 
This is unlike the results from the other datasets where most of the baselines correlate poorly with human judgements. 
A likely explanation is that each data instance from \citet{xu2016optimizing} has eight human references. 
Having multiple human references capture more allowable variations in language quality and therefore, provide a more comprehensive guideline than a single reference. 
In Section \ref{sec:correlation_vs_number_of_references}, we further analyze this phenomenon and discuss how the number of human references affects the results for each evaluation metric.

\subsection{Text Compression}
\label{sec:compression}

\begin{table}[t!]
	\centering
	\small
	\begin{tabular}{cccc}
		\toprule
		& $\rho$ & $\tau$ & $r$ \\
		\midrule
		\textbf{BLEU-best} & 0.21 & 0.15 & 0.21 \\
		\cmidrule{2-4}
		\textbf{ROUGE-best} & 0.41 & 0.29 & 0.41 \\
		\cmidrule{2-4}
		\textbf{METEOR} & 0.33 & 0.23 & 0.32 \\
		\cmidrule{2-4}
		\textbf{TER} & 0.32 & 0.23 & 0.33 \\
		\cmidrule{2-4}
		\textbf{VecSim} & 0.22 & 0.16 & 0.23 \\
		\cmidrule{2-4}
		\textbf{WMD-best} & 0.23 & 0.17 & 0.25 \\
		\cmidrule{2-4}
		\textbf{MoverScore} & 0.34 & 0.24 & 0.34 \\
		\cmidrule{2-4}
		\textbf{SumQE} & 0.38 & 0.23 & 0.43 \\
		\cmidrule{2-4}
		\textbf{\our} & \textbf{0.50} & \textbf{0.37} & \textbf{0.52} \\
		\bottomrule
	\end{tabular}
	\caption{Instance-level Spearman's $\rho$, Kendall's $\tau$ and Pearson's $r$ correlations with \textit{Grammar} on the text compression dataset \cite{toutanova2016dataset}.}
	\label{table:compression}
\end{table}

\textbf{Dataset}: We use the text compression dataset collected in \citet{toutanova2016dataset}. 
It has 2955 instances generated by four machine learning systems and each system output instance receives a human-assigned \textit{Grammar} score.

\noindent \textbf{Results} (Table \ref{table:compression}): We notice that \our outperforms all the other metrics by a significant margin.

\section{Discussion}
\label{sec:analysis}

\begin{figure}
	\centering
	\includegraphics[width=0.49\linewidth]{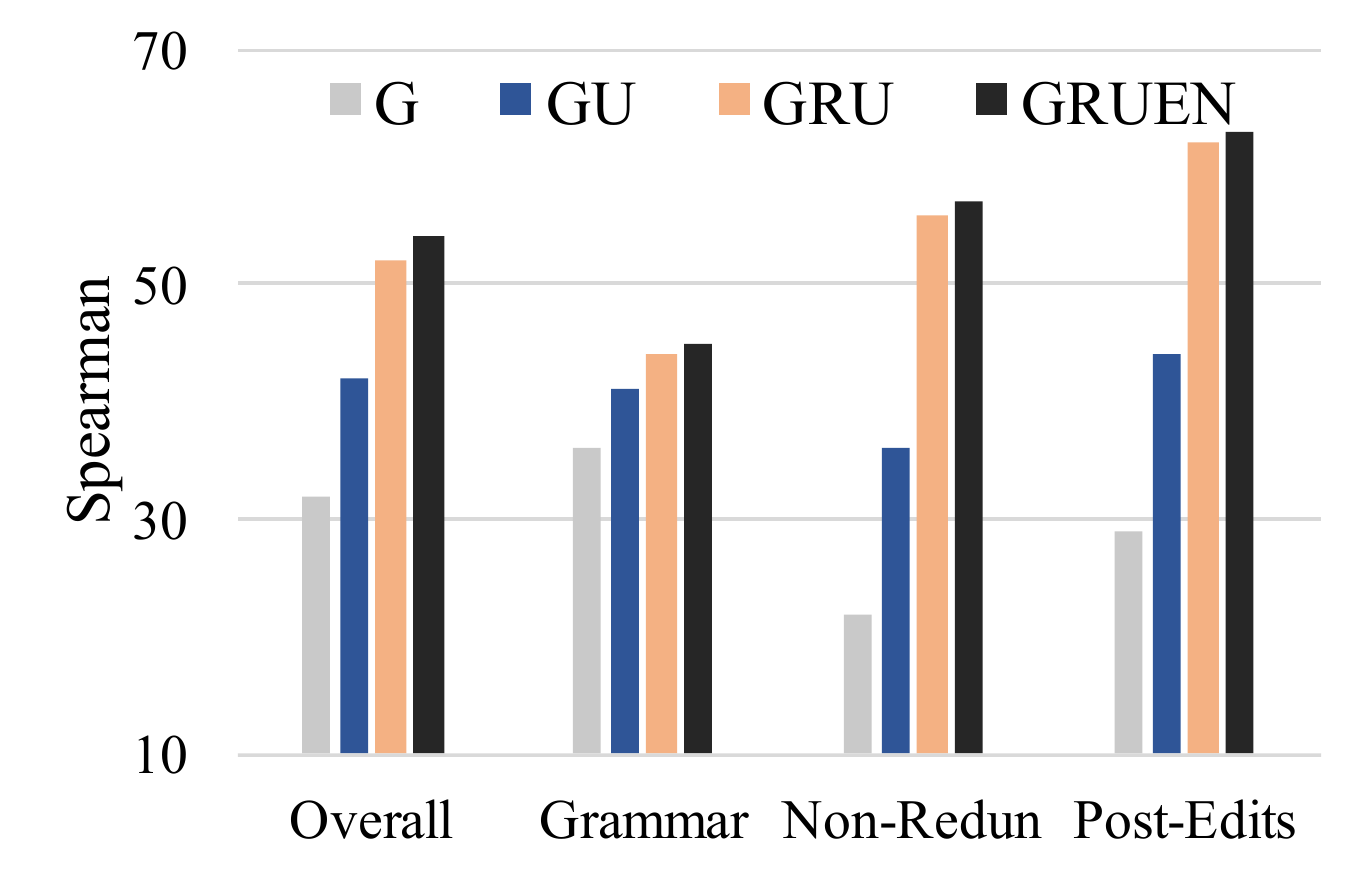}
	\includegraphics[width=0.49\linewidth]{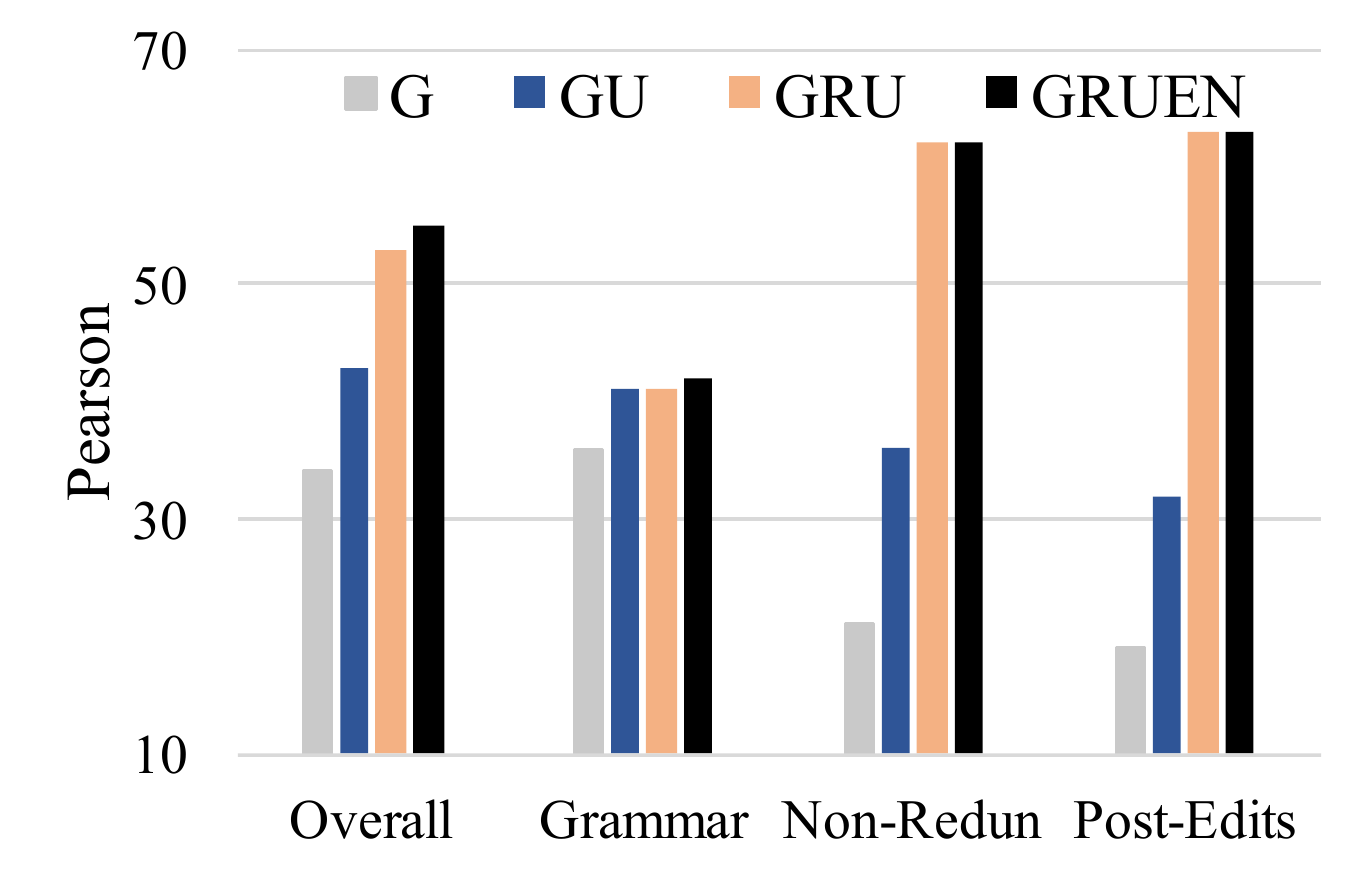}
	\caption{Ablation study on the CNN/Daily Mail Dataset. For better visualization, we present the absolute value of \textit{Post-Edits}.}
	\label{fig:Ablation}
\end{figure}

\begin{figure}
	\centering
	\includegraphics[width=0.495\linewidth]{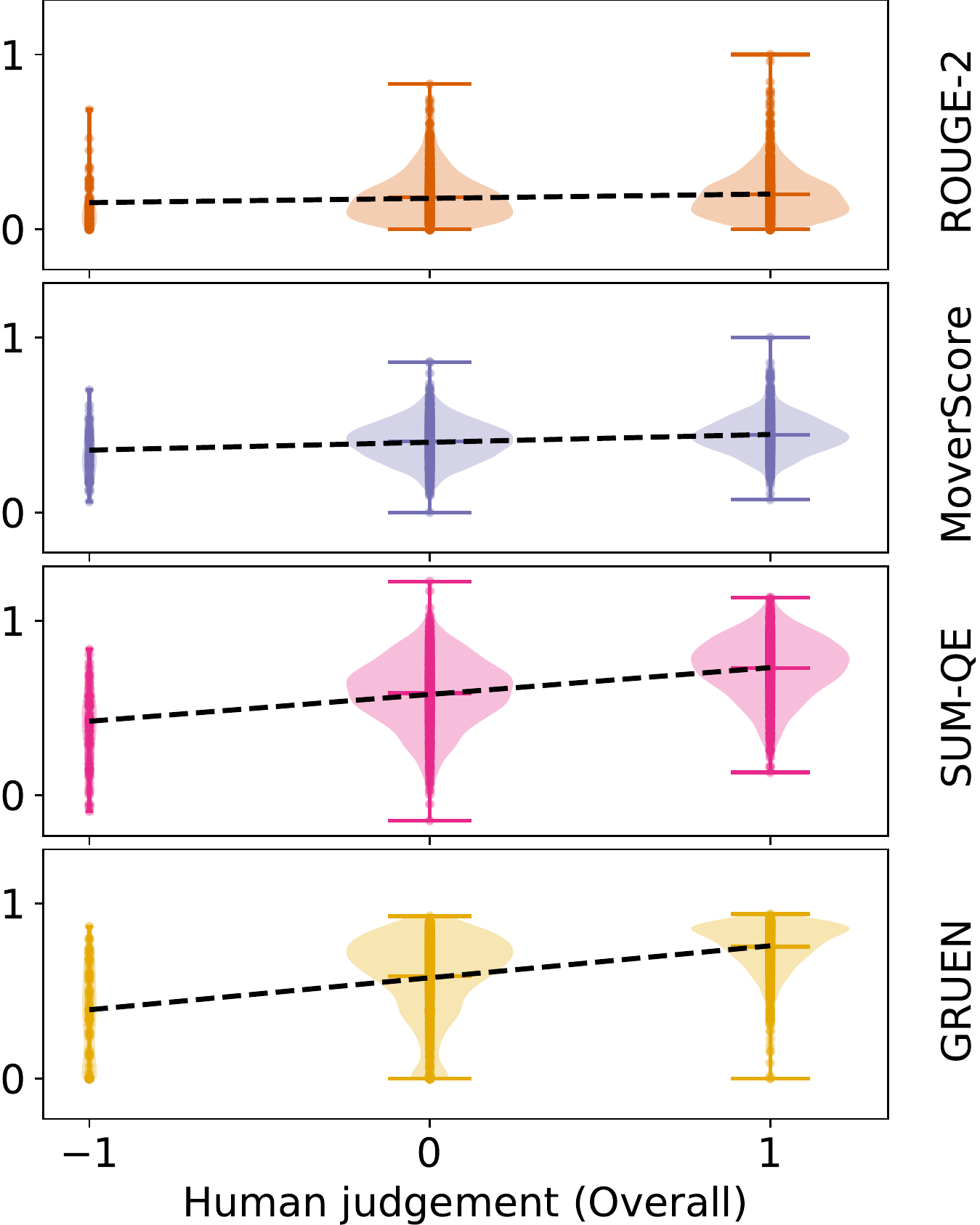}
	\includegraphics[width=0.45\linewidth]{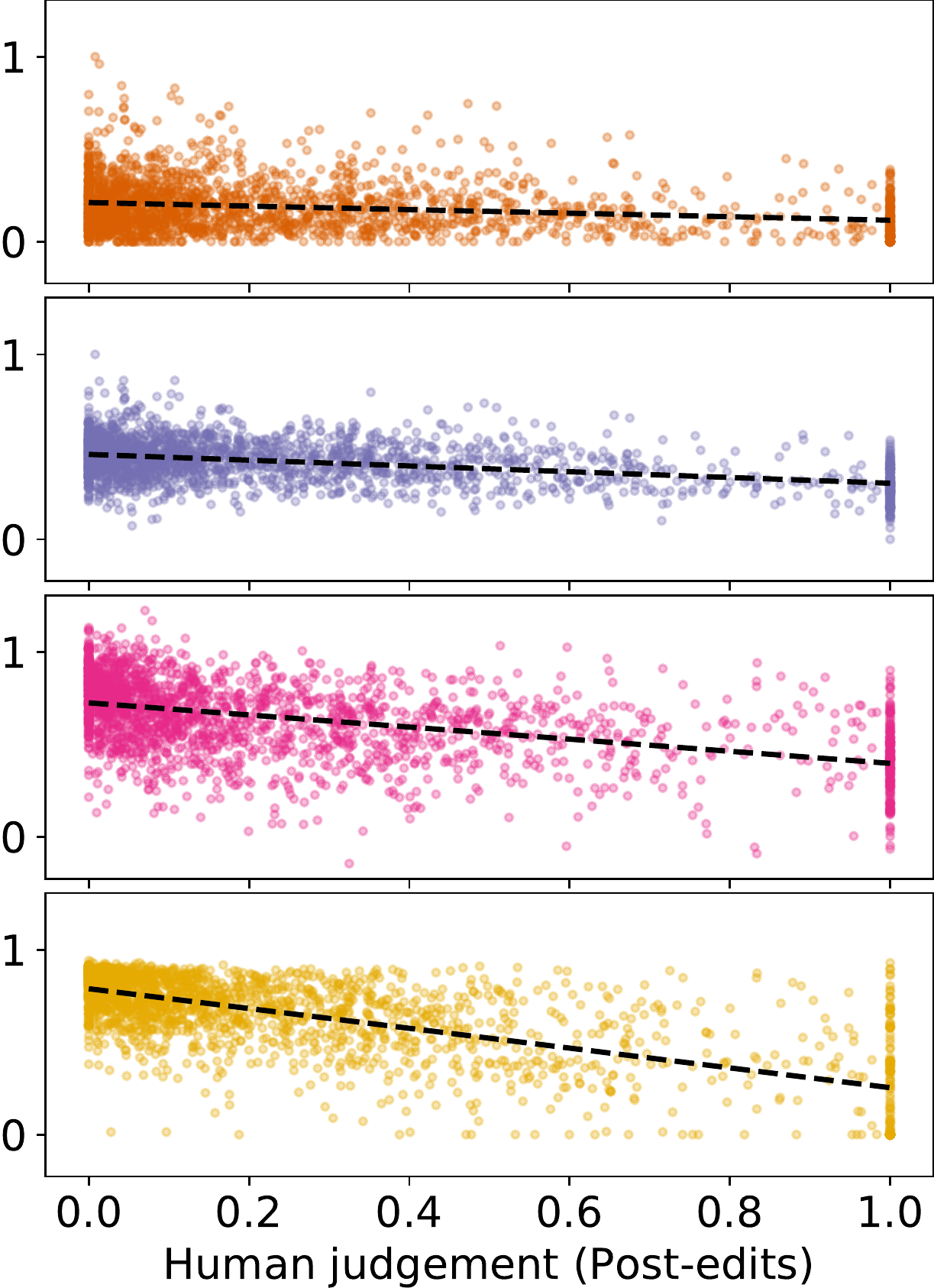}
	\caption{Instance-level distribution of scores for the CNN/Daily Mail dataset. Left shows the \textit{Overall} score distribution on bad (-1), moderate (0) and good (1) outputs. 
		Right shows the scattered \textit{Post-edits} score distribution, which is negatively correlated with the output quality. 
		The dotted line indicates a regression line, which implies the Pearson's correlation $r$.}
	\label{fig:distribution}
\end{figure}

\begin{figure}[ht]
	\centering
	\includegraphics[width=0.48\linewidth]{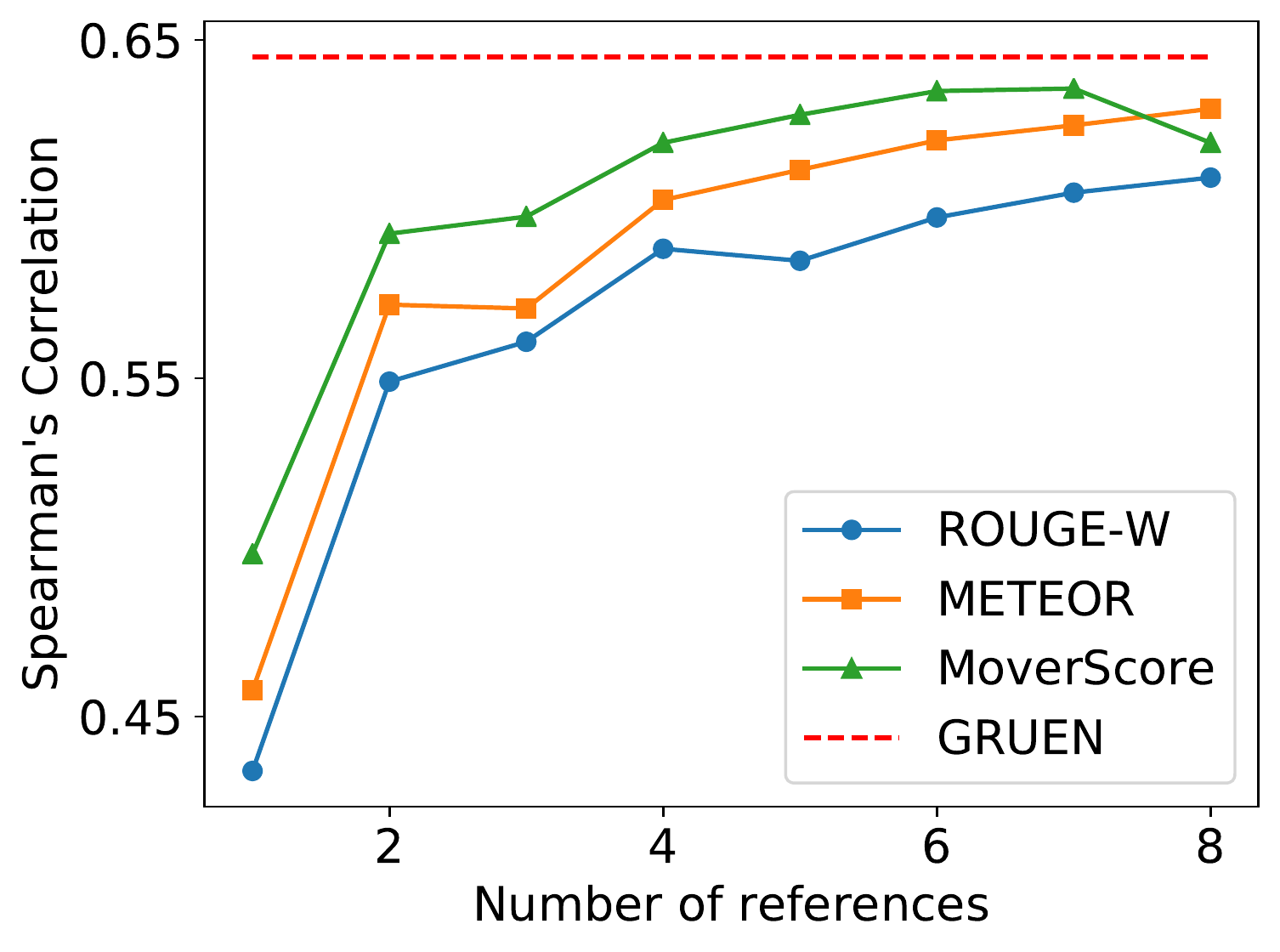}
	\includegraphics[width=0.48\linewidth]{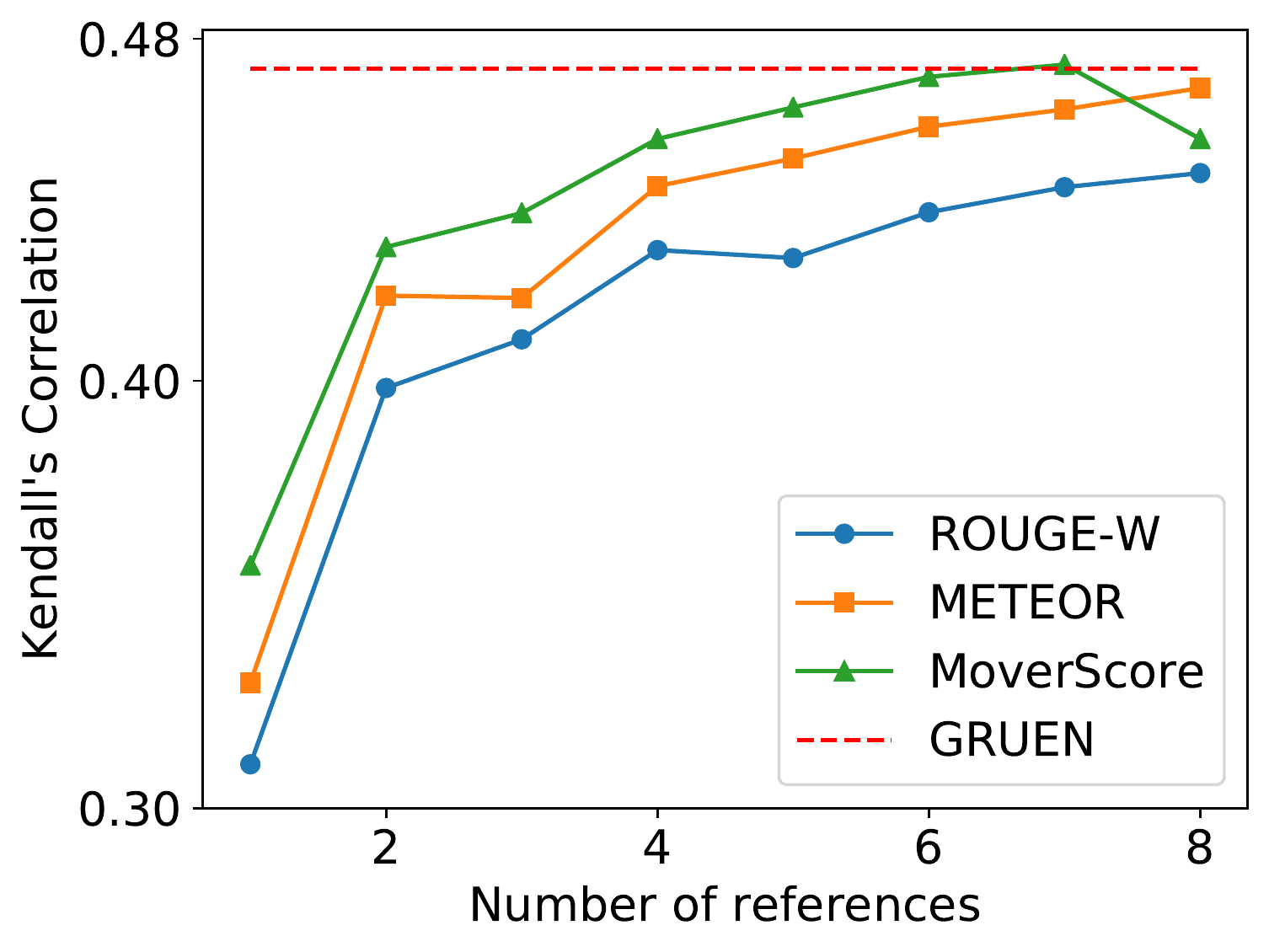}
	\caption{Spearman's Correlation and Kendall's Correlation \textit{v.s.} Number of human references.}
	\label{fig:number_of_references}
\end{figure}

\begin{table*}
	\small
	\centering
	\begin{tabular}{*{3}{c}}
		\toprule
		\textbf{System output examples} & \textbf{Remarks}\\
		\midrule
		\multicolumn{1}{p{0.56\textwidth}}{\textbf{(a) Grammaticality}: Mr Erik Meldik said the. }& 
		\multicolumn{1}{p{0.39\textwidth}}{Incomplete sentence, and hence has low sentence probability and bad grammar score, captured by the BERT language model.} \\
		\midrule
		\multicolumn{1}{p{0.56\textwidth}}{\textbf{(b) Grammaticality}: Orellana shown red card for throwing grass at Sergio Busquets. }& 
		\multicolumn{1}{p{0.39\textwidth}}{Bad grammar captured by the learned knowledge on the CoLA dataset.}\\
		\midrule
		\multicolumn{1}{p{0.56\textwidth}}{\textbf{(c) Non-redundancy}: The brutal murder of Farkhunda, a young woman in Afghanistan, was burnt and callously chucked into a river in Kabul. \underline{The brutal murder of Farkhunda, a young woman in Afghanistan} became pallbearers. }& 
		\multicolumn{1}{p{0.39\textwidth}}{Unnecessary repetition (underlined), which can be avoided by using a pronoun (\ie, she).
			The large overlap between the two sentences is captured by the inter-sentence syntactic features.} \\
		\midrule
		\multicolumn{1}{p{0.56\textwidth}}{\textbf{(d) Focus}: The FDA's Nonprescription Drugs Advisory Committee will meet Oct. Infant cough-and-cold products were approved decades ago without adequate testing in children because experts assumed that children were simply small adults, and that drugs approved for adults must also work in children. Ian Paul, an assistant professor of pediatrics at Penn State College of Medicine who has studied the medicines. } & 
		\multicolumn{1}{p{0.39\textwidth}}{Component sentences are scattered, of different themes or even irrelevant to each other.
			The sentence embedding similarity of each pair of adjacent sentences is low and thus, results in low Focus score.}\\
		\midrule
		\multicolumn{1}{p{0.56\textwidth}}{\textbf{(e) Structure and Coherence}: Firefighters worked with police and ambulance staff to free the boy, whose leg was trapped for more than half an hour down the hole. It is believed the rubber drain cover had been kicked out of position and within hours, the accident occurred. A 12-year-old schoolboy needed to be rescued after falling down an eight-foot drain in Peterborough.} & \multicolumn{1}{p{0.39\textwidth}}{The output is only a heap of related information, where the component sentences are in a unorganized, wrong or incomprehensible order. Its sentence structure and readability can be much improved if the three component sentences are in the order of 3,1,2.}\\
		\bottomrule
	\end{tabular}
	\caption{Case study: linguistic quality analysis}
	\label{tab:caseLQ}
\end{table*}

\begin{table}
	\centering
	\small
	\begin{tabular}{cccc}
		\toprule
		& $\rho$ & $\tau$ & $r$ \\
		\midrule
		\textbf{BLEU-best} & 0.51 & 0.38 & 0.61 \\
		\cmidrule{2-4}
		\textbf{ROUGE-best} & 0.52 & 0.38 & 0.71 \\
		\cmidrule{2-4}
		\textbf{METEOR} & 0.45 & 0.30 & 0.73 \\
		\cmidrule{2-4}
		\textbf{TER} & 0.64 & 0.46 & 0.71 \\
		\cmidrule{2-4}
		\textbf{VecSim} & 0.38 & 0.27 & 0.62 \\
		\cmidrule{2-4}
		\textbf{WMD-best} & 0.31 & 0.23 & 0.60 \\
		\cmidrule{2-4}
		\textbf{MoverScore} & 0.42 & 0.30 & 0.66 \\
		\cmidrule{2-4}
		\textbf{SUM-QE} & 0.76 & 0.63 & 0.69 \\
		\cmidrule{2-4}
		\textbf{\our} & \textbf{0.87} & \textbf{0.69} & \textbf{0.85} \\
		\bottomrule
	\end{tabular}
	\caption{System-level Spearman's $\rho$, Kendall's $\tau$ and Pearson's $r$ correlations with \textit{Readability} on the TAC-2011 dataset.}
	\label{table:system_summarization}
\end{table}

The discussion is primarily conducted for the \textit{text summarization} task considering that \our can measure multiple dimensions in Table \ref{table:DUC_guideline} of the generated text.

\subsection{Ablation study}
The results of the ablation analysis (Figure \ref{fig:Ablation}) show the effectiveness of G (the Grammaticality module alone), GU (the Grammaticality+focUs modules), GRU (the Grammaticality+non-Redundancy+focUs modules) on the summarization output using the CNN/Daily Mail dataset. 
We make the following three observations: 
1) The \textit{Grammar} score is largely accounted for by our grammaticality module, and only marginally  by the others; 
2) The focus and non-redundancy module of \our more directly target the \textit{Post-edits} and \textit{Non-redundancy} aspects of linguistic quality; 
3) The structure and coherence module does not have significant improvement over the linguistic quality dimensions. One possible reason is that structure and coherence is a high-level feature. It is difficult to be captured by not only the models but also the human annotators. 
Please refer to Table \ref{tab:caseLQ} for an example of a system output with poor structure and coherence.

\subsection{Alignment with Rating Scale}
\label{sec:distribution_of_scores}
We compared the scores of ROUGE-2, MoverScore, SUM-QE and \our with those of human judgments on outputs of different quality as shown in Figure \ref{fig:distribution}.
These are in-line with the findings in \citet{chaganty2018price,novikova2017we,zhao2019moverscore} that existing automatic metrics are well correlated with human ratings at the lower end of the rating scale than those in the middle or high end. 
In contrast, we observe that \our is particularly good at distinguishing high-end cases, \ie, system outputs which are rated as good by the human judges.

\subsection{Impact of Number of References}
\label{sec:correlation_vs_number_of_references}
Figure \ref{fig:number_of_references} shows how the Spearman's correlation of each metric varies with different numbers of human references in the text simplification dataset. 
It is clear that existing reference-based metrics show better performance with more human references. 
One possible reason is that the system outputs are compared with more allowable grammatical and semantic variations. 
These allowable variations could potentially make the reference-based metrics better at distinguishing high-end cases, alleviate the shortcoming in Section \ref{sec:distribution_of_scores}, and thus allow the metrics to perform well. 
However, in most cases, it is expensive to collect multiple human references for each instance.

\subsection{Case Study}
Table \ref{tab:caseLQ} presents a case study on examples with poor Grammaticality, Non-redundancy, Focus, and Structure and Coherence. 
In Table~\ref{tab:analysis_redundancy}-\ref{tab:case_AP} in the Appendix, we further analyze how non-redundancy is captured by each of the inter-sentence syntactic features, and also present a comparative study for each linguistic dimension.

\subsection{System-level Correlation}
Our results have shown that \our improves the instance-level correlation performance from poor to moderate. 
At the system-level too, we observe significant improvements in correlation. 
Table \ref{table:system_summarization} shows the system-level linguistic quality correlation scores for \textit{Readability} on the TAC-2011 dataset, which consists of 51 systems (\ie, summarizers). 
At the system level, most baseline metrics have moderate correlations, which aligns with the findings in \citet{chaganty2018price}, while \our achieves a high correlation. 
Note that we do not further study the system-level correlations on other datasets, since they have no more than four systems and thus the correlations are not meaningful to be compared with.

\subsection{Limitations and Future Work}
\label{sec:limitation}
\our evaluates non-redundancy by looking for lexical overlap across sentences. 
However, they still remain unexamined for semantically relevant components that are in different surface forms. 
Besides, it does not handle intra-sentence redundancy, such as ``In 2012, Spain won the European Championships for a second time in 2012.". 
Another challenging problem is to evaluate the referential clarity as proposed in \citet{hoa2006overview}, which is particularly important for long sentences and multi-sentence outputs. 
Future work should aim for a more comprehensive evaluation of redundancy and tackle the referential clarity challenge.

\section{Conclusion}
\label{sec:conclusion}
We proposed \our to evaluate Grammaticality, non-Redundancy, focUs, structure and coherENce of generated text. 
Without requiring human references, \our achieves the new state-of-the-art results on seven datasets over four NLG tasks. 
Besides, as an unsupervised metric, \our is deterministic, free from obtaining costly human judgments, and adaptable to various NLG tasks.

\bibliography{ref}

\begin{thebibliography}{55}
\expandafter\ifx\csname natexlab\endcsname\relax\def\natexlab#1{#1}\fi

\bibitem[{Banerjee and Lavie(2005)}]{banerjee2005meteor}
Satanjeev Banerjee and Alon Lavie. 2005.
\newblock Meteor: An automatic metric for mt evaluation with improved
  correlation with human judgments.
\newblock In \emph{The ACL Workshop on Intrinsic and Extrinsic Evaluation
  Measures for Machine Translation and/or Summarization}.

\bibitem[{Chaganty et~al.(2018)Chaganty, Mussmann, and
  Liang}]{chaganty2018price}
Arun Chaganty, Stephen Mussmann, and Percy Liang. 2018.
\newblock The price of debiasing automatic metrics in natural language
  evalaution.
\newblock In \emph{Association for Computational Linguistics (ACL)}.

\bibitem[{Clark et~al.(2019)Clark, Celikyilmaz, and Smith}]{clark2019sentence}
Elizabeth Clark, Asli Celikyilmaz, and Noah~A Smith. 2019.
\newblock Sentence mover’s similarity: Automatic evaluation for
  multi-sentence texts.
\newblock In \emph{Association for Computational Linguistics (ACL)}.

\bibitem[{Conroy and Dang(2008)}]{conroy2008mind}
John~M Conroy and Hoa~Trang Dang. 2008.
\newblock Mind the gap: Dangers of divorcing evaluations of summary content
  from linguistic quality.
\newblock In \emph{Association for Computational Linguistics (ACL)}.

\bibitem[{Dang(2006)}]{hoa2006overview}
Hoa~Trang Dang. 2006.
\newblock Overview of duc 2006.
\newblock In \emph{Document Understanding Conference (DUC)}.

\bibitem[{Denkowski and Lavie(2014)}]{denkowski2014meteor}
Michael Denkowski and Alon Lavie. 2014.
\newblock Meteor universal: Language specific translation evaluation for any
  target language.
\newblock In \emph{Workshop on Statistical Machine Translation}.

\bibitem[{Devlin et~al.(2019)Devlin, Chang, Lee, and
  Toutanova}]{devlin2019bert}
Jacob Devlin, Ming-Wei Chang, Kenton Lee, and Kristina Toutanova. 2019.
\newblock Bert: Pre-training of deep bidirectional transformers for language
  understanding.
\newblock In \emph{North American Association for Computational Linguistics
  (NAACL)}.

\bibitem[{Doddington(2002)}]{doddington2002automatic}
George Doddington. 2002.
\newblock Automatic evaluation of machine translation quality using n-gram
  co-occurrence statistics.
\newblock In \emph{International Conference on Human Language Technology
  Research}.

\bibitem[{Dorr et~al.(2011)Dorr, Olive, McCary, and
  Christianson}]{dorr2011machine}
Bonnie Dorr, Joseph Olive, John McCary, and Caitlin Christianson. 2011.
\newblock Machine translation evaluation and optimization.
\newblock In \emph{Handbook of Natural Language Processing and Machine
  Translation}, pages 745--843. Springer.

\bibitem[{Dziri et~al.(2019)Dziri, Kamalloo, Mathewson, and
  Zaiane}]{dziri2019evaluating}
Nouha Dziri, Ehsan Kamalloo, Kory~W Mathewson, and Osmar Zaiane. 2019.
\newblock Evaluating coherence in dialogue systems using entailment.
\newblock \emph{arXiv preprint arXiv:1904.03371}.

\bibitem[{Gao et~al.(2020)Gao, Zhao, and Eger}]{gao2020supert}
Yang Gao, Wei Zhao, and Steffen Eger. 2020.
\newblock Supert: Towards new frontiers in unsupervised evaluation metrics for
  multi-document summarization.
\newblock In \emph{Association for Computational Linguistics (ACL)}.

\bibitem[{Graham and Baldwin(2014)}]{graham2014testing}
Yvette Graham and Timothy Baldwin. 2014.
\newblock Testing for significance of increased correlation with human
  judgment.
\newblock In \emph{Empirical Methods in Natural Language Processing (EMNLP)}.

\bibitem[{Graham et~al.(2013)Graham, Baldwin, Moffat, and
  Zobel}]{graham2013continuous}
Yvette Graham, Timothy Baldwin, Alistair Moffat, and Justin Zobel. 2013.
\newblock Continuous measurement scales in human evaluation of machine
  translation.
\newblock In \emph{The Seventh Linguistic Annotation Workshop and
  Interoperability with Discourse}.

\bibitem[{Guo and Hu(2019)}]{guo2019meteor++}
Yinuo Guo and Junfeng Hu. 2019.
\newblock Meteor++ 2.0: Adopt syntactic level paraphrase knowledge into machine
  translation evaluation.
\newblock In \emph{The Fourth Conference on Machine Translation}.

\bibitem[{Hermann et~al.(2015)Hermann, Kocisky, Grefenstette, Espeholt, Kay,
  Suleyman, and Blunsom}]{hermann2015teaching}
Karl~Moritz Hermann, Tomas Kocisky, Edward Grefenstette, Lasse Espeholt, Will
  Kay, Mustafa Suleyman, and Phil Blunsom. 2015.
\newblock Teaching machines to read and comprehend.
\newblock In \emph{Neural Information Processing Systems (NIPS)}.

\bibitem[{Kate et~al.(2010)Kate, Luo, Patwardhan, Franz, Florian, Mooney,
  Roukos, and Welty}]{kate2010learning}
Rohit~J Kate, Xiaoqiang Luo, Siddharth Patwardhan, Martin Franz, Radu Florian,
  Raymond~J Mooney, Salim Roukos, and Chris Welty. 2010.
\newblock Learning to predict readability using diverse linguistic features.
\newblock In \emph{International Conference on Computational Linguistics
  (COLING)}.

\bibitem[{Kiss and Strunk(2006)}]{kiss2006unsupervised}
Tibor Kiss and Jan Strunk. 2006.
\newblock Unsupervised multilingual sentence boundary detection.
\newblock \emph{Computational Linguistics}, 32(4):485--525.

\bibitem[{Knott et~al.(2001)Knott, Oberlander, O’Donnell, and
  Mellish}]{knott2001beyond}
Alistair Knott, Jon Oberlander, Mick O’Donnell, and Chris Mellish. 2001.
\newblock Beyond elaboration: The interaction of relations and focus in
  coherent text.
\newblock \emph{Text Representation: Linguistic and Psycholinguistic Aspects},
  pages 181--196.

\bibitem[{Kusner et~al.(2015)Kusner, Sun, Kolkin, and
  Weinberger}]{kusner2015word}
Matt Kusner, Yu~Sun, Nicholas Kolkin, and Kilian Weinberger. 2015.
\newblock From word embeddings to document distances.
\newblock In \emph{International Conference on Machine Learning (ICML)}.

\bibitem[{Lan et~al.(2019)Lan, Chen, Goodman, Gimpel, Sharma, and
  Soricut}]{lan2019albert}
Zhenzhong Lan, Mingda Chen, Sebastian Goodman, Kevin Gimpel, Piyush Sharma, and
  Radu Soricut. 2019.
\newblock Albert: A lite bert for self-supervised learning of language
  representations.
\newblock \emph{arXiv preprint arXiv:1909.11942}.

\bibitem[{Lavie and Denkowski(2009)}]{lavie2009meteor}
Alon Lavie and Michael~J Denkowski. 2009.
\newblock The meteor metric for automatic evaluation of machine translation.
\newblock \emph{Machine Translation}, 23(2-3):105--115.

\bibitem[{Li et~al.(2016{\natexlab{a}})Li, Galley, Brockett, Gao, and
  Dolan}]{li2016diversity}
Jiwei Li, Michel Galley, Chris Brockett, Jianfeng Gao, and Bill Dolan.
  2016{\natexlab{a}}.
\newblock A diversity-promoting objective function for neural conversation
  models.
\newblock In \emph{North American Association for Computational Linguistics
  (NAACL)}.

\bibitem[{Li et~al.(2016{\natexlab{b}})Li, Monroe, Ritter, Jurafsky, Galley,
  and Gao}]{li2016deep}
Jiwei Li, Will Monroe, Alan Ritter, Dan Jurafsky, Michel Galley, and Jianfeng
  Gao. 2016{\natexlab{b}}.
\newblock Deep reinforcement learning for dialogue generation.
\newblock In \emph{Empirical Methods in Natural Language Processing (EMNLP)}.

\bibitem[{Lin(2004)}]{lin2004rouge}
Chin-Yew Lin. 2004.
\newblock Rouge: A package for automatic evaluation of summaries.
\newblock In \emph{Text Summarization Branches Out}.

\bibitem[{Lin and Hovy(2003)}]{lin2003automatic}
Chin-Yew Lin and Eduard Hovy. 2003.
\newblock Automatic evaluation of summaries using n-gram co-occurrence
  statistics.
\newblock In \emph{North American Association for Computational Linguistics
  (NAACL)}.

\bibitem[{Liu et~al.(2016)Liu, Lowe, Serban, Noseworthy, Charlin, and
  Pineau}]{liu2016not}
Chia-Wei Liu, Ryan Lowe, Iulian Serban, Mike Noseworthy, Laurent Charlin, and
  Joelle Pineau. 2016.
\newblock How not to evaluate your dialogue system: An empirical study of
  unsupervised evaluation metrics for dialogue response generation.
\newblock In \emph{Empirical Methods in Natural Language Processing (EMNLP)}.

\bibitem[{Ma et~al.(2017)Ma, Graham, Wang, and Liu}]{ma2017blend}
Qingsong Ma, Yvette Graham, Shugen Wang, and Qun Liu. 2017.
\newblock Blend: a novel combined mt metric based on direct
  assessment—casict-dcu submission to wmt17 metrics task.
\newblock In \emph{The Second Conference on Machine Translation}.

\bibitem[{Mairesse et~al.(2010)Mairesse, Ga{\v{s}}i{\'c},
  Jur{\v{c}}{\'\i}{\v{c}}ek, Keizer, Thomson, Yu, and
  Young}]{mairesse2010phrase}
Fran{\c{c}}ois Mairesse, Milica Ga{\v{s}}i{\'c}, Filip
  Jur{\v{c}}{\'\i}{\v{c}}ek, Simon Keizer, Blaise Thomson, Kai Yu, and Steve
  Young. 2010.
\newblock Phrase-based statistical language generation using graphical models
  and active learning.
\newblock In \emph{Association for Computational Linguistics (ACL)}.

\bibitem[{Mao et~al.(2020)Mao, Liu, Zhu, Ren, and Han}]{mao2020facet}
Yuning Mao, Liyuan Liu, Qi~Zhu, Xiang Ren, and Jiawei Han. 2020.
\newblock Facet-aware evaluation for extractive summarization.
\newblock In \emph{Association for Computational Linguistics (ACL)}.

\bibitem[{Mikolov et~al.(2013)Mikolov, Sutskever, Chen, Corrado, and
  Dean}]{mikolov2013distributed}
Tomas Mikolov, Ilya Sutskever, Kai Chen, Greg~S Corrado, and Jeff Dean. 2013.
\newblock Distributed representations of words and phrases and their
  compositionality.
\newblock In \emph{Neural Information Processing Systems (NIPS)}.

\bibitem[{Nallapati et~al.(2016)Nallapati, Zhou, dos Santos, glar
  Gul{\c{c}}ehre, and Xiang}]{nallapati2016abstractive}
Ramesh Nallapati, Bowen Zhou, Cicero dos Santos, {\c{C}}a~glar Gul{\c{c}}ehre,
  and Bing Xiang. 2016.
\newblock Abstractive text summarization using sequence-to-sequence rnns and
  beyond.
\newblock \emph{Computational Natural Language Learning (CoNLL)}.

\bibitem[{Nenkova and Passonneau(2004)}]{nenkova2004evaluating}
Ani Nenkova and Rebecca Passonneau. 2004.
\newblock Evaluating content selection in summarization: The pyramid method.
\newblock In \emph{North American Association for Computational Linguistics
  (NAACL)}.

\bibitem[{Ng and Abrecht(2015)}]{ng2015better}
Jun-Ping Ng and Viktoria Abrecht. 2015.
\newblock Better summarization evaluation with word embeddings for rouge.
\newblock In \emph{Empirical Methods in Natural Language Processing (EMNLP)}.

\bibitem[{Novikova et~al.(2017)Novikova, Du{\v{s}}ek, Curry, and
  Rieser}]{novikova2017we}
Jekaterina Novikova, Ond{\v{r}}ej Du{\v{s}}ek, Amanda~Cercas Curry, and Verena
  Rieser. 2017.
\newblock Why we need new evaluation metrics for nlg.
\newblock In \emph{Empirical Methods in Natural Language Processing (EMNLP)}.

\bibitem[{Over et~al.(2007)Over, Dang, and Harman}]{over2007duc}
Paul Over, Hoa Dang, and Donna Harman. 2007.
\newblock Duc in context.
\newblock \emph{Information Processing \& Management}, 43(6):1506--1520.

\bibitem[{Pagliardini et~al.(2018)Pagliardini, Gupta, and
  Jaggi}]{pagliardini2018unsupervised}
Matteo Pagliardini, Prakhar Gupta, and Martin Jaggi. 2018.
\newblock Unsupervised learning of sentence embeddings using compositional
  n-gram features.
\newblock In \emph{North American Association for Computational Linguistics
  (NAACL)}.

\bibitem[{Papineni et~al.(2002)Papineni, Roukos, Ward, and
  Zhu}]{papineni2002bleu}
Kishore Papineni, Salim Roukos, Todd Ward, and Wei-Jing Zhu. 2002.
\newblock Bleu: a method for automatic evaluation of machine translation.
\newblock In \emph{Association for Computational Linguistics (ACL)}.

\bibitem[{Pennington et~al.(2014)Pennington, Socher, and
  Manning}]{pennington2014glove}
Jeffrey Pennington, Richard Socher, and Christopher~D Manning. 2014.
\newblock Glove: Global vectors for word representation.
\newblock In \emph{Empirical Methods in Natural Language Processing (EMNLP)}.

\bibitem[{Pitler et~al.(2010)Pitler, Louis, and Nenkova}]{pitler2010automatic}
Emily Pitler, Annie Louis, and Ani Nenkova. 2010.
\newblock Automatic evaluation of linguistic quality in multi-document
  summarization.
\newblock In \emph{Association for Computational Linguistics (ACL)}.

\bibitem[{ShafieiBavani et~al.(2018)ShafieiBavani, Ebrahimi, Wong, and
  Chen}]{shafieibavani2018graph}
Elaheh ShafieiBavani, Mohammad Ebrahimi, Raymond Wong, and Fang Chen. 2018.
\newblock A graph-theoretic summary evaluation for rouge.
\newblock In \emph{Empirical Methods in Natural Language Processing (EMNLP)}.

\bibitem[{Shimanaka et~al.(2018)Shimanaka, Kajiwara, and
  Komachi}]{shimanaka2018ruse}
Hiroki Shimanaka, Tomoyuki Kajiwara, and Mamoru Komachi. 2018.
\newblock Ruse: Regressor using sentence embeddings for automatic machine
  translation evaluation.
\newblock In \emph{The Third Conference on Machine Translation: Shared Task
  Papers}.

\bibitem[{Snover et~al.(2006)Snover, Dorr, Schwartz, Micciulla, and
  Makhoul}]{snover2006study}
Matthew Snover, Bonnie Dorr, Richard Schwartz, Linnea Micciulla, and John
  Makhoul. 2006.
\newblock A study of translation edit rate with targeted human annotation.
\newblock In \emph{Association for Machine Translation in the Americas}.

\bibitem[{Specia and Shah(2018)}]{specia2018machine}
Lucia Specia and Kashif Shah. 2018.
\newblock Machine translation quality estimation: Applications and future
  perspectives.
\newblock In \emph{Translation Quality Assessment}, pages 201--235. Springer.

\bibitem[{Toutanova et~al.(2016)Toutanova, Brockett, Tran, and
  Amershi}]{toutanova2016dataset}
Kristina Toutanova, Chris Brockett, Ke~M Tran, and Saleema Amershi. 2016.
\newblock A dataset and evaluation metrics for abstractive compression of
  sentences and short paragraphs.
\newblock In \emph{Empirical Methods in Natural Language Processing (EMNLP)}.

\bibitem[{Walker(1998)}]{walker1998centering}
Joshi~Prince Walker. 1998.
\newblock \emph{Centering theory in discourse}.
\newblock Oxford University Press.

\bibitem[{Wang and Cho(2019)}]{wang2019bert}
Alex Wang and Kyunghyun Cho. 2019.
\newblock Bert has a mouth, and it must speak: Bert as a markov random field
  language model.
\newblock In \emph{The Workshop on Methods for Optimizing and Evaluating Neural
  Language Generation}.

\bibitem[{Wang et~al.(2019)Wang, Li, and Smola}]{wang2019language}
Chenguang Wang, Mu~Li, and Alexander~J Smola. 2019.
\newblock Language models with transformers.
\newblock \emph{arXiv preprint arXiv:1904.09408}.

\bibitem[{Warstadt et~al.(2018)Warstadt, Singh, and
  Bowman}]{warstadt2018neural}
Alex Warstadt, Amanpreet Singh, and Samuel~R Bowman. 2018.
\newblock Neural network acceptability judgments.
\newblock \emph{arXiv preprint arXiv:1805.12471}.

\bibitem[{Way(2018)}]{way2018quality}
Andy Way. 2018.
\newblock Quality expectations of machine translation.
\newblock In \emph{Translation Quality Assessment}, pages 159--178. Springer.

\bibitem[{Wen et~al.(2015)Wen, Gasic, Mrk{\v{s}}i{\'c}, Su, Vandyke, and
  Young}]{wen2015semantically}
Tsung-Hsien Wen, Milica Gasic, Nikola Mrk{\v{s}}i{\'c}, Pei-Hao Su, David
  Vandyke, and Steve Young. 2015.
\newblock Semantically conditioned lstm-based natural language generation for
  spoken dialogue systems.
\newblock In \emph{Empirical Methods in Natural Language Processing (EMNLP)}.

\bibitem[{Williams(1959)}]{williams1959regression}
Evan~James Williams. 1959.
\newblock \emph{Regression Analysis}, volume~14.
\newblock Wiley.

\bibitem[{Xenouleas et~al.(2019)Xenouleas, Malakasiotis, Apidianaki, and
  Androutsopoulos}]{xenouleas2019sum}
Stratos Xenouleas, Prodromos Malakasiotis, Marianna Apidianaki, and Ion
  Androutsopoulos. 2019.
\newblock Sum-qe: a bert-based summary quality estimation model.
\newblock In \emph{Empirical Methods in Natural Language Processing and
  International Joint Conference on Natural Language Processing
  (EMNLP-IJCNLP)}.

\bibitem[{Xu et~al.(2016)Xu, Napoles, Pavlick, Chen, and
  Callison-Burch}]{xu2016optimizing}
Wei Xu, Courtney Napoles, Ellie Pavlick, Quanze Chen, and Chris Callison-Burch.
  2016.
\newblock Optimizing statistical machine translation for text simplification.
\newblock \emph{Transactions of the Association for Computational Linguistics
  (TACL)}, 4:401--415.

\bibitem[{Zhang et~al.(2020)Zhang, Kishore, Wu, Weinberger, and
  Artzi}]{zhang2020bertscore}
Tianyi Zhang, Varsha Kishore, Felix Wu, Kilian~Q Weinberger, and Yoav Artzi.
  2020.
\newblock Bertscore: Evaluating text generation with bert.
\newblock In \emph{International Conference on Learning Representations
  (ICLR)}.

\bibitem[{Zhao et~al.(2019)Zhao, Peyrard, Liu, Gao, Meyer, and
  Eger}]{zhao2019moverscore}
Wei Zhao, Maxime Peyrard, Fei Liu, Yang Gao, Christian~M Meyer, and Steffen
  Eger. 2019.
\newblock Moverscore: Text generation evaluating with contextualized embeddings
  and earth mover distance.
\newblock In \emph{Empirical Methods in Natural Language Processing and
  International Joint Conference on Natural Language Processing
  (EMNLP-IJCNLP)}.

\end{thebibliography}
\bibliographystyle{acl_natbib}

\appendix

\section{Quantitative Analysis}
\label{sec:appendix_quantitative}

\begin{table}
	\centering
	\small
	\begin{tabular}{ccccc}
		\toprule
		&  & $\rho$ & $\tau$ & $r$ \\
		\cmidrule{1-5}
		\multirow{4}{*}{\textbf{CNN/}} & \textbf{Overall} & *** & *** & ***  \\
		\cmidrule{2-5}
		\multirow{4}{*}{\textbf{Daily Mail}} &\textbf{Grammar} & * & --- & --- \\
		\cmidrule{2-5}
		&\textbf{Non-Redun} & *** & *** & ***  \\
		\cmidrule{2-5}
		&\textbf{Post-edits} & *** & *** & ***  \\
		\cmidrule{1-5}
		\textbf{TAC-2011} &\textbf{Readability} & * & ** & ** \\
		\cmidrule{1-5}
		\multirow{2}{*}{\textbf{BAGEL}} & \textbf{Naturalness} & 0.01 & 0.07 & ** \\
		\cmidrule{2-5}
		&\textbf{Quality} & 0.06 & 0.17 & * \\
		\cmidrule{1-5}
		\multirow{2}{*}{\textbf{SFHOTEL}} & \textbf{Naturalness} & *** & *** & *** \\
		\cmidrule{2-5}
		&\textbf{Quality} & *** & *** & *** \\
		\cmidrule{1-5}
		\multirow{2}{*}{\textbf{SFREST}} & \textbf{Naturalness} & *** & *** & ***  \\
		\cmidrule{2-5}
		&\textbf{Quality} & *** & *** & ***  \\
		\cmidrule{1-5}
		\multicolumn{1}{p{0.13\textwidth}}{\textbf{\citet{xu2016optimizing}}} &\textbf{Grammar} & 0.33 & 0.46 & --- \\
		\cmidrule{1-5}
		\multicolumn{1}{p{0.13\textwidth}}{\textbf{\citet{toutanova2016dataset}}} &\textbf{Grammar} & *** & *** & *** \\
		\bottomrule
	\end{tabular}
	\caption{William Significance Test on \our against the best baselines for each correlation type and each dataset. *, **, *** indicate the significance level of $<$0.01, $<$0.001 and $<$0.0001 respectively. --- indicates \our does not outperform the best baseline.}
	\label{table:william_significance_test}
\end{table}

\begin{table*}
	\centering
	\small
	\begin{tabular}{cccc|ccc|ccc|ccc}
		\toprule
		& \multicolumn{3}{c}{\textbf{Overall}} & \multicolumn{3}{|c}{\textbf{Grammar}} & \multicolumn{3}{|c}{\textbf{Non-redun}} & \multicolumn{3}{|c}{\textbf{Post-edits}} \\ 
		\cmidrule{2-13} 
		& $\rho$ & $\tau$ & $r$ & $\rho$ & $\tau$ & $r$ & $\rho$ & $\tau$ & $r$ & $\rho$ & $\tau$ & $r$ \\
		\midrule
		\textbf{BLEU-best} & 0.17 & 0.14 & 0.20 & 0.12 & 0.09 & 0.15 & 0.20 & 0.15 & 0.23 & -0.23 & -0.16 & -0.33 \\
		\cmidrule{2-13}
		\textbf{ROUGE-best} & 0.17 & 0.14 & 0.19 & 0.13 & 0.09 & 0.15 & 0.20 & 0.15 & 0.23 & -0.26 & -0.18 & -0.34 \\
		\cmidrule{2-13}
		\textbf{METEOR} & 0.17 & 0.13 & 0.17 & 0.11 & 0.09 & 0.13 & 0.20 & 0.15 & 0.21 & -0.26 & -0.18 & -0.31 \\
		\cmidrule{2-13}
		\textbf{TER} & -0.01 & -0.00 & 0.01 & 0.03 & 0.02 & 0.04 & -0.05 & -0.04 & -0.04 & 0.06 & 0.04 & 0.07 \\
		\cmidrule{2-13}
		\textbf{VecSim} & 0.17 & 0.13 & 0.21 & 0.11 & 0.08 & 0.15 & 0.20 & 0.15 & 0.26 & -0.27 & -0.18 & -0.40 \\
		\cmidrule{2-13}
		\textbf{WMD-best} & 0.27 & 0.21 & 0.26 & 0.24 & 0.18 & 0.25 & 0.27 & 0.20 & 0.25 & -0.31 & -0.21 & -0.29 \\
		\cmidrule{2-13}
		\textbf{MoverScore} & 0.23 & 0.18 & 0.25 & 0.17 & 0.13 & 0.20 & 0.28 & 0.21 & 0.32 & -0.33 & -0.23 & -0.41 \\
		\cmidrule{2-13}
		\textbf{SUM-QE} & 0.53 & 0.43 & 0.54 & 0.49 & \textbf{0.38} & \textbf{0.49} & 0.47 & 0.36 & 0.46 & -0.54 & -0.38 & -0.45 \\
		\cmidrule{2-13}
		\textbf{\our} & \textbf{0.58} & \textbf{0.47} & \textbf{0.58} & \textbf{0.50} & 0.37 & 0.48 & \textbf{0.62} & \textbf{0.48} & \textbf{0.66} & \textbf{-0.68} & \textbf{-0.50} & \textbf{-0.64} \\
		\bottomrule
	\end{tabular}
	\caption{Instance-level Spearman's $\rho$, Kendall's $\tau$ and Pearson's $r$ correlations on the \textbf{reliable} data instances of the CNN/Daily Mail dataset.}
	\label{table:full_reliable}
\end{table*}

\begin{table*}
	\small
	\centering
	\begin{tabular}{*{3}{c}}
		\toprule
		\multicolumn{1}{p{0.8\textwidth}}{\textbf{Example Outputs}} & \multicolumn{1}{p{0.1\textwidth}}{\textbf{Feature}}\\
		\midrule
		\multicolumn{1}{p{0.8\textwidth}}{\textbf{(1)}: The monkey took a bunch of bananas on the desk. It took a bunch of bananas on the desk. }& 
		\multicolumn{1}{p{0.1\textwidth}}{ABCD}\\
		\midrule
		\multicolumn{1}{p{0.8\textwidth}}{\textbf{(2)}: The monkey took a bunch of bananas on the desk. The monkey took a bunch of bananas on the desk, and they are the fruits reserved for the special guests invited tonight. }& 
		\multicolumn{1}{p{0.1\textwidth}}{ABD}\\
		\midrule
		\multicolumn{1}{p{0.8\textwidth}}{\textbf{(3)}: The monkey took a bunch of bananas on the desk. The monkey took a large bunch of bananas on the red desk. }& 
		\multicolumn{1}{p{0.1\textwidth}}{CD}\\
		\midrule
		\multicolumn{1}{p{0.8\textwidth}}{\textbf{(4)}: The monkey took a bunch of bananas on the desk. It took bunches of banana on the desks. }& 
		\multicolumn{1}{p{0.1\textwidth}}{C}\\
		\bottomrule
	\end{tabular}
	\caption{Example with poor non-redundancy. The features that contribute to the non-redundancy penalty are labeled on the right. }
	\label{tab:analysis_redundancy}
\end{table*}

\begin{table*}
	\small
	\centering
	\begin{tabular}{*{3}{c}}
		\toprule
		\textbf{Example Outputs} & \textbf{Module Scores}\\
		\midrule
		\multicolumn{1}{p{0.8\textwidth}}{\textbf{(a) Grammaticality}: Orellana shown red card for throwing grass at Sergio Busquets. }& 
		\multicolumn{1}{p{0.1\textwidth}}{$y_g = 0.2$}\\
		\midrule
		\multicolumn{1}{p{0.8\textwidth}}{\textbf{(b) Grammaticality}: Orellana was shown a red card for throwing grass at Sergio Busquets. }& \multicolumn{1}{p{0.1\textwidth}}{$y_g = 0.7$}\\
		\midrule
		\multicolumn{1}{p{0.8\textwidth}}{\textbf{(c) Non-redundancy}: The brutal murder of Farkhunda, a young woman in Afghanistan, whose body was burnt and callously chucked into a river in Kabul. The brutal murder of Farkhunda, a young woman in Afghanistan became pallbearers, hoisting the victim's coffin on their shoulders draped with headscarves.}& 
		\multicolumn{1}{p{0.1\textwidth}}{$y_r = -0.4$}\\
		\midrule
		\multicolumn{1}{p{0.8\textwidth}}{\textbf{(d) Non-redundancy}: The brutal murder of Farkhunda, a young woman in Afghanistan, whose body was burnt and callously chucked into a river in Kabul. She became pallbearers, hoisting the victim's coffin on their shoulders draped with headscarves.}& 
		\multicolumn{1}{p{0.1\textwidth}}{$y_r = 0.0$}\\
		\midrule
		\multicolumn{1}{p{0.8\textwidth}}{\textbf{(e) Focus}: The FDA's Nonprescription Drugs Advisory Committee will meet Oct. Infant cough-and-cold products were approved decades ago without adequate testing in children because experts assumed that children were simply small adults, and that drugs approved for adults must also work in children. Ian M. Paul, an assistant professor of pediatrics at Penn State College of Medicine who has studied the medicines.} & 
		\multicolumn{1}{p{0.1\textwidth}}{$y_f = -0.1$}\\
		\midrule
		\multicolumn{1}{p{0.8\textwidth}}{\textbf{(f) Focus}: On March 1, 2007, the Food/Drug Administration (FDA) started a broad safety review of children's cough/cold remedies. They are particularly concerned about use of these drugs by infants. By September 28th, the 356-page FDA review urged an outright ban on all such medicines for children under six. Dr. Charles Ganley, a top FDA official said ``We have no data on these agents of what's a safe and effective dose in Children." The review also stated that between 1969 and 2006, 123 children died from taking decongestants and antihistamines. On October 11th, all such infant products were pulled from the markets.} & 
		\multicolumn{1}{p{0.1\textwidth}}{$y_f = 0.0$}\\
		\midrule
		\multicolumn{1}{p{0.8\textwidth}}{\textbf{(g) Coherence and Structure}: Firefighters worked with police and ambulance staff to free the boy, whose leg was trapped for more than half an hour down the hole. It is believed the rubber drain cover had been kicked out of position and within hours, the accident occurred. A 12-year-old schoolboy needed to be rescued after falling down an eight-foot drain in Peterborough.} & \multicolumn{1}{p{0.1\textwidth}}{$y_c = -0.1$}\\
		\midrule
		\multicolumn{1}{p{0.8\textwidth}}{\textbf{(h) Coherence and Structure}: A 12-year-old schoolboy needed to be rescued after falling down an eight-foot drain in Peterborough. Firefighters worked with police and ambulance staff to free the boy, whose leg was trapped for more than half an hour down the hole. It is believed the rubber drain cover had been kicked out of position and within hours, the accident occurred.} & \multicolumn{1}{p{0.1\textwidth}}{$y_c = 0.0$}\\
		\midrule
		\multicolumn{1}{p{0.8\textwidth}}{\textbf{(i) Overall}: The monkey took a bottle of a water bottle in a bid to cool it down with bottle in hand. The monkey is the bottle to its hands before attempting to quench its thirst. It is the the bottle of the bottle in its mouth and a bottle. It's the bottle. A bottle in the water bottle.} & \multicolumn{1}{p{0.1\textwidth}}{$Y_S = 0.0$}\\
		\midrule
		\multicolumn{1}{p{0.8\textwidth}}{\textbf{(j) Overall}: The footage was captured on a warm day in Bali, Indonesia. Tour guide cools monkey down by spraying it with water. Monkey then picks up bottle and casually unscrews the lid. Primate has drink and remarkably spills very little liquid.} & \multicolumn{1}{p{0.1\textwidth}}{$Y_S = 0.8$}\\
		\bottomrule
	\end{tabular}
	\caption{A comparative study on good and bad example outputs for each linguistic aspect.}
	\label{tab:case_AP}
\end{table*}

\subsection{William's Significance Test}
In Table \ref{table:william_significance_test}, we perform William's significance tests on \our against the best baselines for each linguistic score and each correlation measurement (\eg, SUM-QE for $\rho$ on the \textit{Overall} score of the CNN/Daily Mail dataset, METEOR for $r$ on the \textit{Grammar} score of the dataset in \citet{xu2016optimizing}). 
We found that the differences are significant ($p < 0.0001$) in 24 out of 39 cases.

\subsection{Performance on Reliable Instances}
In the human annotation process, each instance receives a score that is the aggregate of multiple people's ratings. Given the subjective nature of the task of annotating for linguistic quality, there are some instances where annotators disagree. 
To analyze how we perform on reliably coded instances, we show in Table \ref{table:full_reliable} the correlation scores on the instances where all annotators agreed perfectly on the \textit{Overall} score for the CNN/Daily Mail dataset ($N=1323$). 
We observe that \our consistently outperforms the baselines on the reliable data instances. 
Importantly, \our and SUM-QE are better correlated with human judgements on the reliable data instances than on all the data instances.

\subsection{Analysis on the Dialogue System Datasets}
\label{sec:app_dialogue}
Table \ref{table:dialogue} has shown an extremely poor correlation with human ratings for the baseline metrics on the BAGEL, SFHOTEL and SFREST datasets. 
\citet{novikova2017we} hypothesizes the reason to be the unbalanced label distribution. 
It turns out that the majority of system outputs are good for \textit{Naturalness} with 64\% and \textit{Quality} (58\%), whereas bad examples are only 7\% in total.\footnote{In a 6-point scale, \textit{bad} comprises low ratings ($\leq$2), while \textit{good} comprises high ratings ($\geq$5).}
Our discussion in Section \ref{sec:distribution_of_scores} further explains the reason. 
Existing metrics are bad at assigning high scores to good outputs and thus, have a very poor correlation in such datasets with mostly good examples. 
In contrast, \our is capable of assigning high scores to good outputs and thus, achieves decent correlation results.

While our correlation results may appear to be slightly different from Table 3 in \citet{novikova2017we}, they are in fact the same. The only difference is the result presentation format. \citet{novikova2017we} presents only the best correlation results for each dataset (\ie, BAGEL, SFHOTEL and SFREST) and each NLG system (\ie, TGEN, LOLS and RNNLG), while we present the average correlation score for each dataset. 
Therefore, in Table 3 of \citet{novikova2017we}, a correlation metric that performs well on one NLG system does not mean it performs equally well on another NLG system. As an example of measuring \textit{Informativeness}, BLEU-1 performs well on the TGEN system for the BAGEL dataset, while it performs poorly on the LOLS system for the BAGEL dataset. 
Therefore, BLEU-1 has only a mediocre correlation score over informativeness for the BAGEL dataset, as presented in our result. 
The analysis in \citet{novikova2017we} is more focused in that it analyzes different metrics in a more restricted manner, whereas our analysis of metrics is more general in that we compare correlation scores regardless of which NLG system the output was generated from.

\section{Qualitative Analysis}
\label{sec:appendix_qualitative}

\subsection{Analysis on Non-redundancy}
\label{sec:analysis_nonredun}
To evaluate the non-redundancy score $y_r$ of a system output, we capture repeated components of a pair of sentences by four empirical \textit{inter-sentence} syntactic features: (A) length of longest common substring, (B) length of longest common words, (C) edit distance, and (D) number of common words. 
Features (A) and (B) focus on continuous word overlap of a pair of sentences.
Intuitively, when most characters of a sentence already appears in the other sentence, the system output should probably have a poor non-redundancy score. 
However, features (A) and (B) fail to make a quality evaluation when the repeated components are of a inflected form (\eg, stemming, lemmatization) or not continuous.
To account for the above limitation, we introduce features (C) and (D) that measures the edit distance and the number of common words respectively. 

To gain more intuition, we present a few examples of poor non-redundancy in Table \ref{tab:analysis_redundancy}.
The features that contribute to the non-redundancy penalty are labeled on the right. 
Case (1) has two almost identical sentences and therefore, captured by all four features. 
However, when the word lengths of the two sentences differ a lot, feature (C) is no longer effective as shown in case (2). 
In case (3) where the word overlap is not continuous (\ie, ``The monkey took a" and ``bunch of bananas on the"), the non-redundancy can only be detected by features (C) and (D). 
In case (4), the composing words are of an inflected form and thus, can not be captured by exact word matching features (\ie, features (A), (B), (D)).
As such, we have the four features to complement each other and aim to capture non-redundancy well.

\subsection{Comparative Study}

Table \ref{tab:case_AP} presents a comparative study on good and bad examples for each linguistic quality aspect, together with their corresponding module scores. 
Besides, we compare two examples with good and bad overall linguistic quality scores.

\section{Complete Results}
\label{sec:Complete_Results}

\begin{table}
	\centering
	\small
	\begin{tabular}{cccc}
		\toprule
		& $\rho$ & $\tau$ & $r$ \\
		\midrule
		\textbf{BLEU-1} & 0.38 & 0.28 & 0.41 \\
		\cmidrule{2-4}
		\textbf{BLEU-2} & 0.47 & 0.33 & 0.49 \\
		\cmidrule{2-4}
		\textbf{BLEU-3} & 0.52 & 0.37 & 0.55 \\
		\cmidrule{2-4}
		\textbf{BLEU-4} & 0.55 & 0.40 & 0.58 \\
		\cmidrule{2-4}
		\textbf{ROUGE-1} & 0.51 & 0.37 & 0.56 \\
		\cmidrule{2-4}
		\textbf{ROUGE-2} & 0.54 & 0.39 & 0.58 \\
		\cmidrule{2-4}
		\textbf{ROUGE-3} & 0.52 & 0.38 & 0.55 \\
		\cmidrule{2-4}
		\textbf{ROUGE-4} & 0.50 & 0.36 & 0.51 \\
		\cmidrule{2-4}
		\textbf{ROUGE-L} & 0.56 & 0.40 & 0.59 \\
		\cmidrule{2-4}
		\textbf{ROUGE-W} & 0.61 & 0.45 & 0.64 \\
		\cmidrule{2-4}
		\textbf{WMD} & 0.43 & 0.31 & 0.33 \\
		\cmidrule{2-4}
		\textbf{SMD} & 0.30 & 0.21 & 0.30 \\
		\cmidrule{2-4}
		\textbf{S+WMD} & 0.40 & 0.29 & 0.34 \\
		\bottomrule
	\end{tabular}
	\caption{Instance-level Spearman's $\rho$, Kendall's $\tau$ and Pearson's $r$ correlations with \textit{Grammar} on the text simplification dataset \cite{xu2016optimizing}.}
	\label{table:full_simplification}
\end{table}

\begin{table}
	\centering
	\small
	\begin{tabular}{cccc}
		\toprule
		& $\rho$ & $\tau$ & $r$ \\
		\midrule
		\textbf{BLEU-1} & 0.07 & 0.05 & 0.17 \\
		\cmidrule{2-4}
		\textbf{BLEU-2} & 0.12 & 0.08 & 0.18 \\
		\cmidrule{2-4}
		\textbf{BLEU-3} & 0.17 & 0.12 & 0.19 \\
		\cmidrule{2-4}
		\textbf{BLEU-4} & 0.21 & 0.15 & 0.21 \\
		\cmidrule{2-4}
		\textbf{ROUGE-1} & 0.21 & 0.15 & 0.24 \\
		\cmidrule{2-4}
		\textbf{ROUGE-2} & 0.33 & 0.24 & 0.34 \\
		\cmidrule{2-4}
		\textbf{ROUGE-3} & 0.35 & 0.26 & 0.37 \\
		\cmidrule{2-4}
		\textbf{ROUGE-4} & 0.35 & 0.25 & 0.36 \\
		\cmidrule{2-4}
		\textbf{ROUGE-L} & 0.39 & 0.28 & 0.37 \\
		\cmidrule{2-4}
		\textbf{ROUGE-W} & 0.41 & 0.29 & 0.41 \\
		\cmidrule{2-4}
		\textbf{WMD} & 0.18 & 0.13 & 0.16 \\
		\cmidrule{2-4}
		\textbf{SMD} & 0.23 & 0.17 & 0.25 \\
		\cmidrule{2-4}
		\textbf{S+WMD} & 0.20 & 0.14 & 0.21 \\
		\bottomrule
	\end{tabular}
	\caption{Instance-level Spearman's $\rho$, Kendall's $\tau$ and Pearson's $r$ correlations with \textit{Grammar} on the text compression dataset \cite{toutanova2016dataset}.}
	\label{table:full_compression}
\end{table}

\begin{table*}
	\centering
	\small
	\begin{tabular}{ccc|cc|cc|cc|cc}
		\toprule
		& \multicolumn{8}{c}{\textbf{CNN/Daily Mail}} & \multicolumn{2}{|c}{\textbf{TAC-2011}} \\ 
		\cmidrule{2-11}
		& \multicolumn{2}{c}{\textbf{Overall}} & \multicolumn{2}{|c}{\textbf{Grammar}} & \multicolumn{2}{|c}{\textbf{Non-redun}} & \multicolumn{2}{|c}{\textbf{Post-edits}} & \multicolumn{2}{|c}{\textbf{Readability}} \\ 
		\cmidrule{2-11} 
		& $\rho$  & $r$ & $\rho$ &  $r$ & $\rho$ &  $r$ & $\rho$ & $r$ & $\rho$  & $r$ \\
		\midrule
		\textbf{BLEU-1} & 0.07 & 0.08 & 0.05 & 0.05 & 0.06 & 0.06 & -0.08 & -0.10 & 0.17 & 0.34 \\
		\cmidrule{2-11}
		\textbf{BLEU-2} & 0.13 & 0.14 & 0.09 & 0.09 & 0.13 & 0.14 & -0.16 & -0.20 & 0.21 & 0.35 \\
		\cmidrule{2-11}
		\textbf{BLEU-3} & 0.16 & 0.18 & 0.10 & 0.12 & 0.17 & 0.19 & -0.21 & -0.27 & 0.24 & 0.36 \\
		\cmidrule{2-11}
		\textbf{BLEU-4} & 0.17 & 0.18 & 0.11 & 0.12 & 0.17 & 0.20 & -0.21 & -0.29 & 0.26 & 0.38 \\
		\cmidrule{2-11}
		\textbf{ROUGE-1} & 0.17 & 0.19 & 0.11 & 0.13 & 0.20 & 0.23 & -0.24 & -0.32 & 0.25 & 0.36 \\
		\cmidrule{2-11}
		\textbf{ROUGE-2} & 0.14 & 0.13 & 0.09 & 0.10 & 0.15 & 0.15 & -0.18 & -0.21 & 0.25 & 0.26 \\
		\cmidrule{2-11}
		\textbf{ROUGE-3} & 0.12 & 0.10 & 0.08 & 0.09 & 0.13 & 0.11 & -0.16 & -0.16 & 0.24 & 0.19 \\
		\cmidrule{2-11}
		\textbf{ROUGE-4} & 0.10 & 0.08 & 0.08 & 0.08 & 0.11 & 0.09 & -0.14 & -0.13 & 0.20 & 0.15 \\
		\cmidrule{2-11}
		\textbf{ROUGE-L} & 0.12 & 0.13 & 0.10 & 0.12 & 0.11 & 0.12 & -0.17 & -0.19 & 0.25 & 0.36 \\
		\cmidrule{2-11}
		\textbf{ROUGE-W} & 0.14 & 0.14 & 0.10 & 0.12 & 0.13 & 0.14 & -0.18 & -0.19 & 0.26 & 0.34 \\
		\cmidrule{2-11}
		\textbf{WMD} & 0.18 & 0.11 & 0.12 & 0.10 & 0.19 & 0.11 & -0.23 & -0.15 & 0.19 & 0.17 \\
		\cmidrule{2-11}
		\textbf{SMD} & 0.26 & 0.24 & 0.20 & 0.21 & 0.26 & 0.23 & -0.29 & -0.26 & 0.15 & 0.25 \\
		\cmidrule{2-11}
		\textbf{S+WMD} & 0.21 & 0.17 & 0.15 & 0.15 & 0.22 & 0.17 & -0.26 & -0.21 & 0.19 & 0.24 \\
		\bottomrule
	\end{tabular}
	\caption{Instance-level Spearman's $\rho$ and Pearson's $r$ correlations on the CNN/Daily Mail and TAC-2011 datasets.}
	\label{table:full_summarization}
\end{table*}

\begin{table*}
	\centering
	\small
	\begin{tabular}{ccc|cc|cc|cc|cc|cc}
		\toprule
		& \multicolumn{4}{c}{\textbf{BAGEL}} & \multicolumn{4}{|c}{\textbf{SFHOTEL}} & \multicolumn{4}{|c}{\textbf{SFREST}}\\ 
		\cmidrule{2-13}
		& \multicolumn{2}{c}{\textbf{Naturalness}} & \multicolumn{2}{|c}{\textbf{Quality}} & \multicolumn{2}{|c}{\textbf{Naturalness}} & \multicolumn{2}{|c}{\textbf{Quality}} & \multicolumn{2}{|c}{\textbf{Naturalness}} & \multicolumn{2}{|c}{\textbf{Quality}} \\ 
		\cmidrule{2-13} 
		& $\rho$  & $r$ & $\rho$  & $r$ & $\rho$ & $r$ & $\rho$ & $r$ & $\rho$ & $r$ & $\rho$  & $r$\\
		\midrule
		\textbf{BLEU-1} & -0.02 & -0.02 & -0.02 & -0.01 & 0.03 & 0.11 & -0.04 & 0.04 & 0.03 & 0.03 & -0.03 & -0.02 \\
		\cmidrule{2-13}
		\textbf{BLEU-2} & 0.00 & 0.00 & -0.01 & 0.01 & 0.01 & 0.09 & -0.08 & 0.00 & 0.03 & 0.02 & -0.03 & -0.03 \\
		\cmidrule{2-13}
		\textbf{BLEU-3} & 0.01 & 0.03 & 0.01 & 0.03 & 0.00 & 0.08 & -0.10 & -0.01 & 0.03 & 0.02 & -0.03 & -0.03 \\
		\cmidrule{2-13}
		\textbf{BLEU-4} & 0.03 & 0.04 & 0.02 & 0.05 & 0.00 & 0.07 & -0.10 & -0.02 & 0.03 & 0.03 & -0.03 & -0.02 \\
		\cmidrule{2-13}
		\textbf{ROUGE-1} & 0.10 & 0.12 & 0.10 & 0.12 & -0.01 & 0.06 & -0.11 & -0.03 & 0.02 & 0.01 & -0.05 & -0.04 \\
		\cmidrule{2-13}
		\textbf{ROUGE-2} & 0.11 & 0.13 & 0.10 & 0.12 & -0.02 & 0.02 & -0.12 & -0.07 & 0.02 & 0.03 & -0.06 & -0.04 \\
		\cmidrule{2-13}
		\textbf{ROUGE-3} & 0.08 & 0.10 & 0.07 & 0.09 & -0.03 & 0.01 & -0.12 & -0.06 & 0.01 & 0.04 & -0.06 & -0.03 \\
		\cmidrule{2-13}
		\textbf{ROUGE-4} & 0.04 & 0.09 & 0.04 & 0.08 & -0.04 & 0.00 & -0.12 & -0.06 & 0.02 & 0.05 & -0.04 & -0.01 \\
		\cmidrule{2-13}
		\textbf{ROUGE-L} & 0.08 & 0.10 & 0.09 & 0.11 & -0.01 & 0.07 & -0.11 & -0.03 & 0.01 & 0.01 & -0.06 & -0.04 \\
		\cmidrule{2-13}
		\textbf{ROUGE-W} & 0.08 & 0.10 & 0.08 & 0.10 & -0.02 & 0.04 & -0.12 & -0.05 & 0.05 & 0.05 & -0.03 & -0.02 \\
		\cmidrule{2-13}
		\textbf{WMD} & 0.03 & 0.05 & 0.05 & 0.08 & -0.02 & 0.00 & -0.12 & -0.07 & 0.03 & 0.05 & -0.05 & 0.00 \\
		\cmidrule{2-13}
		\textbf{SMD} & 0.00 & 0.04 & 0.02 & 0.07 & 0.00 & 0.01 & -0.09 & -0.06 & -0.01 & 0.03 & -0.07 & -0.01 \\
		\cmidrule{2-13}
		\textbf{S+WMD} & 0.02 & 0.05 & 0.04 & 0.08 & -0.01 & 0.00 & -0.11 & -0.07 & 0.02 & 0.05 & -0.06 & -0.01 \\
		\bottomrule
	\end{tabular}
	\caption{Instance-level Spearman's $\rho$ and Pearson's $r$ correlations on the BAGEL, SFHOTEL and SFREST datasets.}
	\label{table:full_dialogue}
\end{table*}

We present the complete results of BLEU, ROUGE and WMD for all tasks in Table \ref{table:full_simplification}-\ref{table:full_dialogue}. 

\end{document}